\documentclass[12pt]{iopart}

\usepackage{url}
\usepackage{hyperref}
\usepackage{cite}
\usepackage{algorithm,algorithmic}
\usepackage{color}
\usepackage{graphicx}
\usepackage{iopams}
\usepackage[whole]{bxcjkjatype}  

\DeclareRobustCommand{\bm}[1]{\mathbf{\boldsymbol{#1}}}
\DeclareRobustCommand{\supp}[1]{{\rm supp}{(#1)}}

\DeclareRobustCommand{\diag}[1]{\mathop{\rm Diagm}(#1)}

\renewcommand{\a}{s}  
\renewcommand{\b}{t}  

\newcommand{\1}{\mbox{1}\hspace{-0.25em}\mbox{l}}

\begin{document}

\title[
    Semi-analytic approximate stability selection for correlated data in generalized linear models
]{
    Semi-analytic approximate stability selection for correlated data in generalized linear models
}

\author{Takashi Takahashi and Yoshiyuki Kabashima
    \footnote{Present address: Institute for Physics of Intelligence \& Department of Physics Graduate School of Science, The University of Tokyo
    7-3-1 Hongo, Bunkyo-ku, Tokyo 113-0033, Japan}
}
\ead{takahashi.t.cc@m.titech.ac.jp}
\address{Department of Mathematical and Computing Science \\
Tokyo Institute of Technology\\
2-12-1, Ookayama, Meguro-ku, Tokyo, Japan}

\begin{abstract}
We consider the variable selection problem of generalized linear models (GLMs). Stability selection (SS) is a promising method proposed for solving this problem. Although SS provides practical variable selection criteria, it is computationally demanding because it needs to fit GLMs to many re-sampled datasets. We propose a novel approximate inference algorithm that can conduct SS without the repeated fitting. The algorithm is based on the replica method of statistical mechanics and vector approximate message passing of information theory. For datasets characterized by rotation-invariant matrix ensembles,  we derive state evolution equations that macroscopically describe the dynamics of the proposed algorithm. We also show that their fixed points are consistent with the replica symmetric solution obtained by the replica method. Numerical experiments indicate that the algorithm exhibits fast convergence and high approximation accuracy for both synthetic and real-world data.
\end{abstract}

%

\tableofcontents

\section{Introduction}
\label{sec:introduction}
Modern statistics require the handling of high-dimensional data. The term \emph{high-dimensional} refers to the situation where the ratio of the number of measurements and the number of the parameters is of order 1. Among the many tasks in high-dimensional statistics, variable selection of statistical models is a notoriously difficult problem. In high-dimensional settings, standard sparse regression methods, including the least absolute shrinkage and selection operator (LASSO) method \cite{tibshirani1996regression}, suffer from the problem of choosing the regularization parameter. Although re-sampling methods, such as stability selection (SS) \cite{meinshausen2010ss}, can provide much more accurate variable selection criteria, these methods require substantial computational costs.

As an example, let us consider variable selection in logistic regression. In this regression, we have a dataset $D=\{(\bm{a}_\mu, y_\mu)\}_{\mu=1}^M$, where each $\bm{a}_\mu=(a_{\mu1},a_{\mu2},\dots, a_{\mu N})^\top\in\mathbb{R}^N$ is an $N$-dimensional vector of features or predictors, and each $y_\mu\in\{-1,1\}$ is the associated binary response variable. We denote by $\top$ the matrix/vector transpose. The response variables are independently generated based on a true parameter $\bm{x}_0=(x_{0,1},x_{0,2},\dots,x_{0,N})^\top \in\mathbb{R}^N$ as
\begin{equation}
    y_\mu \sim \frac{1}{1+e^{-\bm{a}_\mu^\top\bm{x}_0}}\delta(y_\mu - 1) + \frac{1}{1+e^{\bm{a}_\mu^\top \bm{x}_0}}\delta(y_\mu + 1),\, \mu=1,2,\dots,M.
    \label{eq:likelihood}
\end{equation}
We denote by $\supp{\bm{x}_0} = \{i \mid x_{0,i}\neq 0, i=1,2,...,N\}$ the support of $\bm{x}_0$. The goal of variable selection is to estimate $\supp{\bm{x}_0}$ from the dataset $D$. In high-dimensional settings, a simple strategy is to use $\ell_1$ regularized logistic regression or LASSO \cite{tibshirani1996regression}. LASSO seeks an estimator of $\bm{x}_0$ as
\begin{equation}
    \hat{\bm{x}}(\gamma,D) = \arg\min_{\bm{x}\in\mathbb{R}^N}\left[-\sum_{\mu=1}^M\log\frac{1}{1+e^{-y_\mu\bm{a}_\mu^\top \bm{x}}} + \gamma\sum_{i=1}^N|x_i|\right],
    \label{eq:lasso}
\end{equation}
where $\gamma>0$ is a parameter that controls the strength of the $\ell_1$ regularizer. The $\ell_1$ regularization term $\gamma\sum_{i=1}^N |x_i|$ allows LASSO to select variables by shrinking a part of the estimated parameters exactly to $0$. For any given regularization parameter $\gamma$, LASSO estimates $\supp{\bm{x}_0}$ as 
\begin{equation}
    \hat{S}(\gamma,D) \equiv \left\{i \mid \hat{x}_i(\gamma,D)\neq 0, i=1,2,\dots, N\right\}.
\end{equation}
Unfortunately, this estimated support $\hat{S}(\gamma,D)$ depends strongly on the choice of the regularization parameter $\gamma$ in real-world datasets. Hence, choosing the regularization parameter for variable selection can be more challenging than for prediction of the response variable where cross-validation is guaranteed to offer the optimal choice on average if features are generated independently from an identical distribution \cite{homrighausen2014leave}.

SS was proposed for tackling this difficulty. We denote by $D^\ast=\{(\bm{a}_1^\ast, y_1^\ast), (\bm{a}_2^\ast, y_2^\ast)$, $\dots, (\bm{a}_M^\ast, y_M^\ast)\}$ a resampled dataset of size $M$ drawn with replacement from $D$. For this resampled dataset, the resampling probability $\Pi_i(\gamma)$ that the variable $i$ is included in the estimated support is given by
\begin{equation}
    \Pi_i(\gamma) = {\rm Prob}_{D^\ast}\left[\hat{x}_i(\gamma,D^\ast)\neq0\right].
    \label{eq:active_prob}
\end{equation}
The probability in (\ref{eq:active_prob}) is with respect to the random resampling and it equals the relative frequency for $\hat{x}_i(\gamma, D^\ast)\neq0$ over all $M^M$ resampled dataset with size $M$. The probability in (\ref{eq:active_prob}) can be approximated by $B$ random samples $D_1^\ast, D_2^\ast, \dots, D_B^\ast$ ($B$ should be large):
\begin{equation}
    \Pi_i(\gamma) \simeq \frac{1}{B}\sum_{b=1}^B \1\left(\hat{x}_i(\gamma, D_b^\ast)\neq 0\right),
\end{equation}
where $\1(...)$ is the indicator function. This probability is termed the \emph{selection probability} and measures the \emph{stability} of each variable. SS chooses variables that have large selection probabilities. The original literature \cite{meinshausen2010ss} combined the above resampling procedure with the randomization of the regularization parameter $\gamma$ as follows
\begin{eqnarray}
    \Pi_i(\gamma_0) &= {\rm Prob}_{D^\ast, \bm{\gamma}}\left[\hat{x}_i(\bm{\gamma}, D^\ast)\neq 0\right], \, i=1,2,\dots,N, 
    \label{eq:active_prob_ss}
    \\
    \hat{\bm{x}}(\bm{\gamma}, D^\ast) &= \arg\min_{\bm{x}\in\mathbb{R}^N}\left[-\sum_{\mu=1}^M \log \frac{1}{1+e^{-y_\mu^\ast(\bm{a}_\mu^\ast)^\top \bm{x}}} + \sum_{i=1}^N\gamma_i|x_i|\right],
    \label{eq:lasso_ss}
    \\
    \gamma_i &\sim \frac{1}{2}\delta(\gamma_i - \gamma_0) + \frac{1}{2}\delta(\gamma_i - 2\gamma_0), i=1,2,\dots,N. \label{eq:dist_regularization}
\end{eqnarray}

\begin{figure}[t]
     \centerline{
        \includegraphics[width=\linewidth]{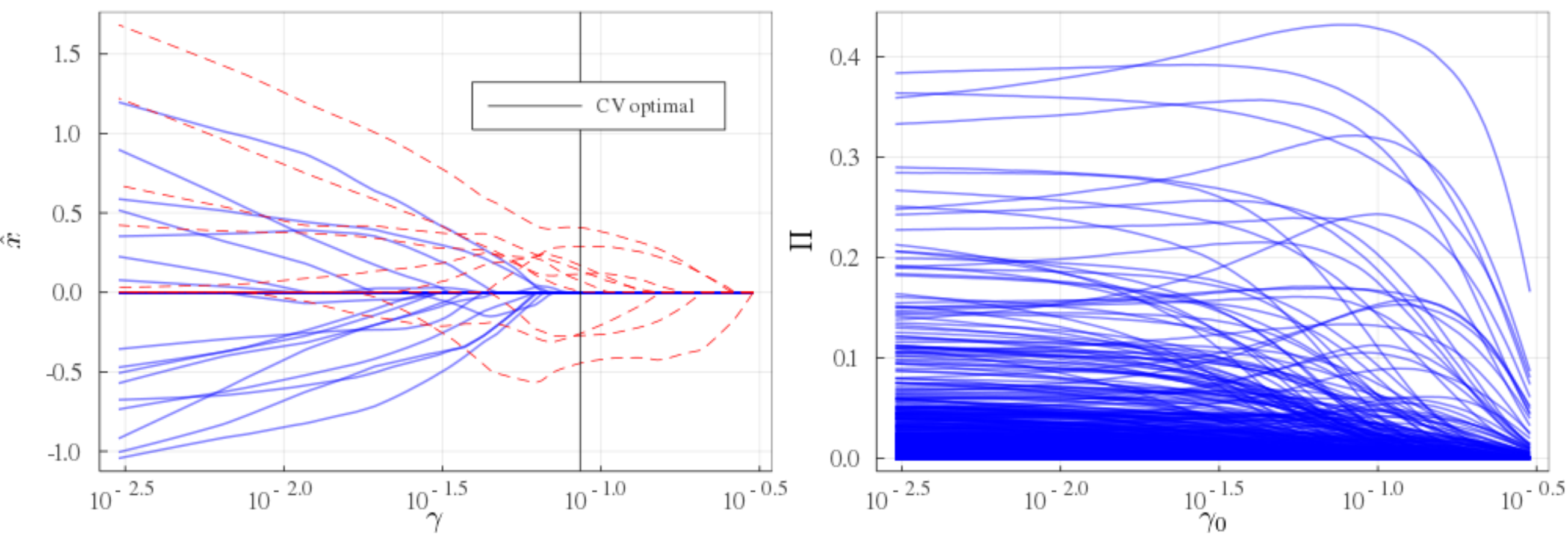}
     }
     \caption{
        Left: The LASSO solutions $\hat{\bm{x}}(\gamma, D)$ based on (\ref{eq:lasso}) for the colon cancer dataset with $M=62$ and $N=2000$. The vertical line corresponds to the cross-validation optimal regularization parameter. The red-dashed lines represent variables chosen by the cross-validation procedure. The non-zero variables strongly depend on the choice of the regularization parameter $\gamma$. 
        Right: The selection probability $\bm{\Pi}(\lambda_0)$ based on (\ref{eq:active_prob_ss}). The selection probability is less dependent on the choice of $\gamma_0$, indicating that choosing the regularization parameter is less critical than the naive LASSO.
    }
    \label{fig:compare_path_stability}
\end{figure}

Figure \ref{fig:compare_path_stability} illustrates the comparison of the LASSO solution (\ref{eq:lasso}) and the selection probability (\ref{eq:active_prob_ss}). Here we used the colon cancer dataset \cite{alon1999broad}. The task is to distinguish cancer from normal tissue using the micro-array data with $N=2000$ features per example. The data were obtained from $22$ normal ($y_\mu=-1$) and $40$ ($y_\mu=1$) cancer tissues. The total number of the samples is $M=62$. The left panel of figure \ref{fig:compare_path_stability} shows the LASSO solutions for the various regularization parameters. Non-zero variables depend strongly on $\gamma$. Choosing the proper value of $\gamma$ is difficult for the original LASSO. Although the cross-validation can optimize the prediction for the response variable, this choice often includes false positive elements \cite{buhlmann2011statistics}. The right panel of figure \ref{fig:compare_path_stability} shows the selection probability for various $\gamma_0$ in (\ref{eq:dist_regularization}). This figure motivates that choosing the regularization parameter $\gamma_0$ is much less critical for the selection probability and that the selection probability approach has a better chance of selecting truly relevant variables.

A major drawback of SS is its computational cost. SS repeatedly solves the $\ell_1$ regularized logistic regression in (\ref{eq:lasso_ss}) for multiple resampled datasets and regularization parameters. The number of resampled datasets and regularization parameters $B$ needs to be large so that the selection probability is reliably estimated. 

In this study, we address the problem of this computational cost. We propose a novel approximate inference algorithm that can conduct SS without repeated fitting. The algorithm is based on the replica method \cite{mezard1987spin} of statistical mechanics and \emph{vector approximate message passing} (VAMP) \cite{schniter2016vector, rangan2019vector} of information theory. We term our algorithm \emph{replicated VAMP} (rVAMP).

The rest of the paper is organized as follows. In section \ref{sec:setup}, we describe stability selection in generalized linear models (GLMs) that we will focus on, and in section \ref{sec:rVAMP}, we derive the proposed algorithm using the replica method and VAMP. In section \ref{sec:macro analysis}, we analyze the proposed algorithm in a large system limit under the assumption that the set of features is characterized by rotation-invariant matrix ensembles. There, we derive the state evolution for self-averaging rVAMP that macroscopically describes the convergence dynamics of rVAMP in an approximate manner, and show that its fixed point is consistent with the replica symmetric solution. In section \ref{sec:numerical experiment}, we apply the proposed algorithm to logistic regression. Through numerical experiments, we confirm the validity of our theoretical analysis and show that the proposed algorithm exhibits fast convergence and high approximation accuracy for both synthetic and real-world data. The final section is devoted to a summary and conclusion.

\subsection{Related work}
Malzahn and Opper first proposed a combination of the replica method and approximate inference to reduce the computational cost of resampling methods \cite{malzahn2003approximate, malzahn2003statistical, malzahn2004approximate}. They demonstrated that employing the adaptive Thouless-Anderson-Palmer (TAP) method\cite{opper2001tractable, opper2001adaptive}, as an approximate inference algorithm, can accurately estimate the bootstrap generalization error for Gaussian process classification/regression. However, the poor convergence of this method is a major flaw of their approach. The adaptive TAP method is based on a naive iteration of TAP equations. The literature in information theory has revealed that the convergence property of such naive iteration scheme is terribly bad \cite{rangan2019vector, bolthausen2014iterative, cakmak2014s}. Thus it requires to find a correct choice of initial conditions. As an algorithm, the adaptive TAP method is undesirable because approximate inference aims to save computation time.

The aforementioned algorithmic problem has been significantly improved by the discovery of approximate message passing (AMP) algorithms in information theory. This type of algorithms was first introduced as an efficient signal processing algorithm \cite{kabashima2003cdma}. \cite{kabashima2003cdma} analyzed its convergence dynamics in a large system limit and showed its fast convergence. \cite{kabashima2003cdma} also revealed that the fixed point of the AMP algorithm shares the same fixed point with the corresponding TAP equation, and thus, AMP can be used as an efficient algorithm to solve the TAP equation. Subsequently, \cite{donoho2009message, bolthausen2014iterative} developed its mathematically rigorous analysis. These rigorous analyses were further generalized in \cite{bayati2011dynamics, javanmard2013state}. However, the above analyses are based on the assumptions that the elements of the feature vectors are independently and identically distributed (i.i.d.) zero-mean random variables, which is not realistic in the context of statistics. To go beyond such simple distributions, VAMP and similar generalizations \cite{schniter2016vector, rangan2019vector, ma2017orthogonal} were developed based on expectation propagation (EP) of machine learning \cite{minka2001expectation, opper2005expectation}.  Under the assumption that feature matrices, whose rows are composed of each feature vectors, are drawn from rotation-invariant random matrix ensembles, VAMP algorithms were analyzed in a large system limit. These analyses derived the convergence dynamics of the VAMP algorithms and revealed that their fixed points are consistent with the corresponding adaptive TAP equations \cite{schniter2016vector, rangan2019vector, takahashi2020macroscopic, ccakmak2018expectation, ccakmak2019convergent, ccakmak2019memory}. In this paper, we extend such VAMP algorithms to replicated systems for approximately performing SS in GLMs.

\cite{obuchi2019semi} proposed an AMP-based approximate resampling algorithm for SS. However, the algorithm assumes independence between the features and was developed for linear regression only. A preliminary application of VAMP to SS in linear regression was also demonstrated \cite{takahashi2019replicated}. In the present study, we further generalize the use of VAMP to GLMs, and also carry out a theoretical analysis of this method. 

\subsection{Notations}
Here we introduce some shorthand notations used throughout the paper.
We denote by $[\omega_i]_{1\le i\le N}$ a vector $\bm{\omega}=(\omega_1,\omega_2,\dots,\omega_N)^\top \in\mathbb{R}^N$. Similarly, we denote by $[\Omega_{\mu i}]_{1\le \mu \le M \atop 1\le i\le N}$ an $M\times N$ matrix whose $\mu i$-th entry is $\Omega_{\mu i}$. For an integer $n=1,2,\dots$, 
we denote by $\bm{1}_n=(1,1,\dots,1)\in\mathbb{R}^n$ a constant vector. For integers $n\in\mathbb{N},m\in\mathbb{Z}$, and vectors $\bm{\omega}=[\omega_i]_{1\le i\le n}, \bm{\psi}=[\psi_i]_{1\le i\le n}$, we denote by $\bm{\omega}/\bm{\psi} = [\omega_i/\psi_i]_{1\le i\le n}$ and $\bm{\omega}^m=[\omega_i^m]_{1\le i\le n}$ component-wise operations. Finally, $\langle \bm{\omega}\rangle\equiv\sum_{i=1}^n \omega_i/n$.

\section{Stability selection in generalized linear models}
\label{sec:setup}
In the following, we consider SS in generalized linear regression/classification. We have a dataset $D=\{(\bm{a}_\mu, y_\mu)\}_{\mu=1}^M$, where each $\bm{a}_\mu=(a_{\mu1}, a_{\mu2}, \dots, a_{\mu N})^\top \in \mathbb{R}^N$ is an $N$-dimensional vector of features or predictors, and each $y_\mu\in\mathcal{Y} \subset \mathbb{R}$ is the associated response variable. The domain of the response variables $\mathcal{Y}$ includes $\mathbb{R}$ for regression and $\{-1,1\}$ for classification. We also use matrix/vector notations $A=[a_{\mu i}]_{1\le \mu \le M \atop 1\le i \le N}\in\mathbb{R}^{M\times N}$ and $\bm{y} = (y_1, y_2,\dots, y_M)^\top\in\mathcal{Y}^M$.

Let $D^\ast=\{(\bm{a}_1^\ast, y_1^\ast), \dots, (\bm{a}_M^\ast, y_M^\ast)\}$ be a resampled dataset composed of $M$ data points drawn with replacement from $D$. Some data point $(\bm{a}_\mu, y_\mu)$ in $D$ appears multiple times in $D^\ast$, and while others do not appear at all.  SS in generalized linear regression/classification computes the selection probability $\bm{\Pi}\in [0,1]^N$ by repeatedly refitting GLMs $p_{y | z}$ for multiple resampled datasets and regularization parameters:
\begin{eqnarray}
    \Pi_i(\gamma_0) &= {\rm Prob}_{D^\ast, \bm{\gamma}}\left[\hat{x}_i(\bm{\gamma}, D^\ast)\neq 0\right],  \,  i=1,2,\dots, N,
    \label{eq:ss_glm}
    \\
    \hat{\bm{x}}(\bm{\gamma}, D^\ast) &= \arg\min_{\bm{x}\in\mathbb{R}^N}\left[-\sum_{\mu=1}^M \log p_{y | z}(y_\mu^\ast | (\bm{a}_\mu^\ast)^\top \bm{x}) + \sum_{i=1}^N\gamma_i |x_i| \right], 
    \label{eq:lasso_ss_glm}
    \\
    \gamma_i &\sim \frac{1}{2}\delta(\gamma_i - \gamma_0) + \frac{1}{2}\delta(\gamma_i - 2\gamma_0), \, i=1,2,...,N, 
    \label{eq:dist_gamma_ss_glm}
\end{eqnarray}
where $\gamma_0>0$ is a control parameter that determines the amount of the regularization. The goal of this paper is to develop a computationally efficient algorithm that returns $\bm{\Pi}(\gamma_0)$ for any positive $\gamma_0$.

\section{Replicated vector approximate message passing}
\label{sec:rVAMP}
To approximate the computation of the selection probability $\bm{\Pi}$, we will use the replica method and VAMP. This section provides a derivation of the proposed algorithm.

\subsection{Occupation vector representation of sampling with replacement}
For convenience, let us introduce the occupation vector representation of the resampled dataset $D^\ast$. The resampled dataset $D^\ast$ is composed of $M$ data points sampled from $D$ with replacement. Hence, it can be represented by a vector of \emph{occupation} numbers $\bm{c}=(c_1,c_2,\dots, c_M)^\top  \in \{0,1,\dots,M\}^M$ with $\sum_{\mu=1}^M c_\mu=M$, where $c_\mu$ is the number of times that the data point $(\bm{a}_\mu, y_\mu)$ appears in $D^\ast$. Although the strict distribution of $\bm{c}$ is the multinomial distribution, for large $M$, the correlation among $\{c_\mu\}_{\mu=1}^M$ is weak. By ignoring this correlation, we can approximate the distribution of $\bm{c}$ by a product of Poisson distribution with mean $1$ \cite{malzahn2003approximate} as:
\begin{equation}
    p(\bm{c}) \simeq \prod_{\mu=1}^M \frac{e^{-1}}{c_\mu!}.
    \label{eq:poisson}
\end{equation}
In this way, we can rewrite the average with respect to $D^\ast$ by the average over the random variable $\bm{c}\in\{0,1,\dots\}^M$ that follows the probability distribution (\ref{eq:poisson}), which is simple and easy to handle.

\subsection{Statistical mechanical formulation of stability selection}
The selection probability $\bm{\Pi}$ in (\ref{eq:ss_glm}) is defined through the optimization problem in (\ref{eq:lasso_ss_glm}). To use techniques of statistical mechanics and approximate inference algorithm, we introduce the Boltzmann distribution as 
\begin{eqnarray}
\fl
    p^{(\beta)}(\bm{x}, \bm{z}; \bm{c}, \bm{\gamma}, D) &= \frac{1}{Z^{(\beta)}(\bm{c}, \bm{\gamma}, D)}\delta(\bm{z} - A\bm{x})\prod_{\mu=1}^M p_{y | z}(y_\mu | z_\mu)^{\beta c_\mu} \prod_{i=1}^N e^{-\beta \gamma_i |x_i|} , 
    \label{eq:boltzmann_dist}
    \\
\fl
    Z^{(\beta)}(\bm{c}, \bm{\gamma}, D) &= \int \delta(\bm{z} - A\bm{x})\prod_{\mu=1}^M p_{y | z}(y_\mu | z_\mu)^{\beta c_\mu} \prod_{i=1}^N e^{-\beta \gamma_i |x_i|}  d\bm{x}d\bm{z}, 
    \label{eq:partition_function}
\end{eqnarray}
where $\bm{x}\in\mathbb{R}^N$, $\bm{z}\in\mathbb{R}^M$, $\beta>0$ is the inverse temperature, and $Z$ is the partition function. The random variables $\bm{\gamma}$ and $\bm{c}$ follow distributions (\ref{eq:dist_gamma_ss_glm}) and (\ref{eq:poisson}), respectively. Then the selection probability can be written using the Boltzmann distribution at the zero-temperature limit as follows:
\begin{eqnarray}
    \Pi_i(\gamma_0) &= \mathbb{E}_{\bm{c},\bm{\gamma}}\left[\1(\hat{x}_i(\bm{c}, \bm{\gamma})\neq 0)\right], \, i=1,2,\dots, N, 
    \\
    \hat{x}_i(\bm{c}, \bm{\gamma}) &= \lim_{\beta\to\infty} \int x_i p^{(\beta)}(\bm{x}, \bm{z}; \bm{c}, \bm{\gamma}, D) d\bm{x}d\bm{z}.
    \label{eq:marginal_moment}
\end{eqnarray}
In the rest of the paper, we will omit the argument $D$ when there is no risk of confusion to avoid cumbersome notation. Still, note that we calculate the above quantities only for the \emph{fixed} dataset $D$.

\subsection{Replica method for semi-analytic approximate resampling method} 
\label{subsec:replica}
Our purpose is to compute the selection probability $\bm{\Pi}(\gamma_0)$ for any $\gamma_0>0$. For this, we compute the distribution of $\hat{x}_i$:
\begin{equation}
    p(m_i) = \mathbb{E}_{\bm{c}, \bm{\gamma}}\left[
        \1\left(
                m_i - \hat{x}_i\left(\bm{c},\bm{\gamma}\right)
            \right)
    \right],
    \label{eq:marginal_dist}
\end{equation}
which is reduced to computing the moments $\mathbb{E}_{\bm{c}, \bm{\gamma}}[\hat{x}_i^r(\bm{c}, \bm{\gamma})]$ for any $r=1,2,\dots$. We now describe how the replica method can be used for this purpose, following the approach of \cite{malzahn2003approximate}.

We use $d^r\bm{x} = d\bm{x}_1d\bm{x}_2\dots d\bm{x}_r$ to denote a measure over $\mathbb{R}^{N\times r}$, with $\bm{x}_1 = (x_{1,1},\dots, x_{1,N})^\top, \dots, \bm{x}_r = (x_{r,1}, \dots, x_{r,N})^\top$.
Analogously, we denote by $d^r\bm{z} = d\bm{z}_1d\bm{z}_2\dots d\bm{z}_r$ as a measure over $\mathbb{R}^{M\times r}$, with $\bm{z}_1=(z_{1,1}, \dots, z_{1,M})^\top, \dots, \bm{z}_r=(z_{r,1}, \dots, z_{r,M})^\top$. Using the definition (\ref{eq:marginal_moment}), the moments $\mathbb{E}_{\bm{c}, \bm{\gamma}}[\hat{x}_i^r(\bm{c}, \bm{\gamma})]$ can be formally written as\footnote{
    Since the aim of this paper is not to provide rigorous analysis, we assume that the exchange of limits, integrals, etc, such as $\mathbb{E}_{\bm{c},\bm{\gamma}}[\lim_{\beta\to\infty}\dots] =\lim_{\beta\to\infty}\mathbb{E}_{\bm{c},\bm{\gamma}}[\dots]$, are possible throughout the paper without further justification.
}
\begin{eqnarray}
\fl
    \mathbb{E}_{\bm{c}, \bm{\gamma}}\left[
        \hat{x}_i^{r}(\bm{c}, \bm{\gamma}) 
    \right]
    &= \lim_{\beta\to\infty}\mathbb{E}_{\bm{c}, \bm{\gamma}}\left[
        \int \prod_{\a=1}^r x_{\a,i} 
        \prod_{\a=1}^r p^{(\beta)}(\bm{x}_{\a}, \bm{z}_{\a})
        d^r\bm{x}d^r\bm{z}
    \right]
    \nonumber 
    \\
\fl
    &= \lim_{\beta\to\infty}\int 
            \prod_{\a=1}^r x_{\a, i} 
            \mathbb{E}_{\bm{c}, \bm{\gamma}}\left[
                \prod_{\a=1}^r 
                \left\{
                    \frac{1}{Z^{(\beta)}(\bm{c}, \bm{\gamma})}
                    \delta(\bm{z}_{\a} - A\bm{x}_{\a})
                \right.
            \right.
            \nonumber \\
            &\hspace{3cm}
            \left.
                \left.
                    \times\prod_{\mu=1}^M p_{y | z}(y_\mu | z_{\a,\mu})^{\beta c_\mu} \prod_{i=1}^N e^{-\beta \gamma_i |x_{\a,i}|}
                \right\}
            \right]
        d^r\bm{x}d^r\bm{z},
        \label{eq:moment_raw}
\end{eqnarray}
which is difficult to evaluate analytically due to the presence of the partition function that depends on $\bm{c}$ and $\bm{\gamma}$ in the denominator. The replica trick \cite{mezard1987spin} bypasses this problem via an identity $\lim_{n\to0}Z^{n-r}=Z^{-r}$. Using this identity, (\ref{eq:moment_raw}) is formally re-expressed as 
\begin{equation}
    \mathbb{E}_{\bm{c}, \bm{\gamma}}\left[
        \hat{x}_i^{r}(\bm{c}, \bm{\gamma}) 
    \right]
    = \lim_{n\to0}\lim_{\beta\to\infty}\mathcal{A}_{i,n}^{(\beta)},
\end{equation}
where 
\begin{eqnarray}
    \mathcal{A}_{i,n}^{(\beta)} = \int 
            &\prod_{\a=1}^r x_{\a, i} 
            \mathbb{E}_{\bm{c}, \bm{\gamma}}\left[
                \left(
                    Z^{(\beta)}(\bm{c}, \bm{\gamma})
                \right)^{n-r}
                \prod_{\a=1}^r 
                \Biggl\{
                    \delta(\bm{z}_{\a} - A\bm{x}_{\a})
            \right.
            \nonumber \\
            &\left.
                \left.
                \times\prod_{\mu=1}^M p_{y | z}(y_\mu | z_{\a,\mu})^{\beta c_\mu} \prod_{i=1}^N e^{-\beta \gamma_i |x_{\a,i}|}
                \right\}
            \right]
        d^r\bm{x}d^r\bm{z}.
\end{eqnarray}
The advantage of this formula is that for integers $n \ge r$, the negative power of the partition function $(Z^{(\beta)}(\bm{c}, \bm{\gamma}))^{-r}$ is eliminated by an integral with respect to $n$ replicated variables. More precisely, using the definition of the partition function (\ref{eq:partition_function}), we obtain
\begin{eqnarray}
\fl
    \mathcal{A}_{i,n}^{(\beta)} = \Xi_n \int 
        \prod_{\a=1}^r x_{\a,i}
        \frac{1}{\Xi_n}
            \prod_{\a=1}^n \delta(\bm{z}_{\a} - A\bm{x}_{\a})
            \prod_{\mu=1}^M \mathbb{E}_{c_\mu}\left[
                \prod_{\a=1}^n p_{y | z}(y_\mu | z_{\a,\mu})^{\beta c_\mu}
            \right]
            \nonumber \\
            \times
            \prod_{i=1}^N \mathbb{E}_{\gamma_i}\left[
                \prod_{\a=1}^n e^{-\beta \gamma_i |x_{\a,i}|}
            \right]
        d^n\bm{x}d^n\bm{z},
        \label{eq:replica expression}
\end{eqnarray}
where $\Xi_n$ is the normalization constant 
\begin{equation}
\fl
    \Xi_n = \int 
        \prod_{\a=1}^n \delta(\bm{z}_{\a} - A\bm{x}_{\a})
            \prod_{\mu=1}^M \mathbb{E}_{c_\mu}\left[
                \prod_{\a=1}^n p_{y | z}(y_\mu | z_{\a,\mu})^{\beta c_\mu}
            \right]
            \prod_{i=1}^N \mathbb{E}_{\gamma_i}\left[
                \prod_{\a=1}^n e^{-\beta \gamma_i |x_{\a,i}|}
            \right]
    d^n\bm{x}d^n\bm{z}.
\end{equation}
The expression (\ref{eq:replica expression}) is much easier to evaluate than the negative power of the partition function.
We call the probability density function given by
\begin{eqnarray}
\fl
    p^{(\beta)}(\{\bm{x}_{\a}\}_{\a=1}^n, \{\bm{z}_{\a}\}_{\a=1}^n) 
    = \frac{1}{\Xi_n}
    \prod_{\a=1}^n \delta(\bm{z}_{\a} - A\bm{x}_{\a})
    \nonumber \\
            \times
            \prod_{\mu=1}^M \mathbb{E}_{c_\mu}\left[
                \prod_{\a=1}^n p_{y | z}(y_\mu | z_{\a,\mu})^{\beta c_\mu}
            \right]
            \prod_{i=1}^N \mathbb{E}_{\gamma_i}\left[
                \prod_{\a=1}^n e^{-\beta \gamma_i |x_{\a,i}|}
            \right],
    \label{eq:replicated system}
\end{eqnarray}
the replicated system. Note that by construction $\lim_{n\to0}\Xi_n=1$.

In this way, we have replaced the original problem with computing first moments of the replicated system (\ref{eq:replicated system}). Of course, we wouldn't expect that we could compute the moments exactly. Otherwise we should have obtained the exact solution without using the replicas. The replica method evaluates a formal expression of $\lim_{\beta\to\infty}\mathcal{A}_{i,n}^{(\beta)}$ for $n=r+1,r+2,\dots$ under appropriate approximations, and then extrapolates it as $n\to0$.

To obtain a formal expression of $\lim_{\beta\to\infty}\mathcal{A}_{i,n}^{(\beta)}$, the following observation is critical. Because the replicated system (\ref{eq:replicated system}) is merely a product of the $n$-copied systems, it is intrinsically invariant under any permutations of $\{(\bm{x}_1, \bm{z}_1), (\bm{x}_2, \bm{z}_2), \dots, (\bm{x}_n, \bm{z}_n)\}$. This property is termed the replica symmetry. From this property, de Finetti's representation theorem \cite{hewitt1955symmetric} guarantees that the replicated system (\ref{eq:replicated system}) is expressed as 
\begin{equation}
    p^{(\beta)}(\{\bm{x}_{\a}\}_{\a=1}^n, \{\bm{z}_{\a}\}_{\a=1}^n)= \int \prod_{\a=1}^n p^{(\beta)}(\bm{x}_{\a}, \bm{z}_{\a} | \bm{\eta}) p^{(\beta)}(\bm{\eta})d\bm{\eta},
    \label{eq:de Finetti}
\end{equation}
where $\bm{\eta}$ is a vector of some random variables that reflects the effects of $\bm{c}$ and $\bm{\gamma}$. This expression indicates that $\mathcal{A}_{i,n}^{(\beta)}$ is reduced to a considerably simple form 
\begin{eqnarray}
\fl
    \mathcal{A}_{i,n}^{(\beta)} &= \int \left(\int x_i p^{(\beta)}(\bm{x}, \bm{z}|\bm{\eta})d\bm{x}d\bm{z}\right)^r \left(\int p^{(\beta)}(\bm{x}, \bm{z}|\bm{\eta})d\bm{x}d\bm{z}\right)^{n-r} p^{(\beta)}(\bm{\eta})d\bm{\eta}
    \nonumber \\
\fl    
    &= \int \left(\int x_i p^{(\beta)}(\bm{x}, \bm{z}|\bm{\eta})d\bm{x}d\bm{z}\right)^r p^{(\beta)}(\bm{\eta})d\bm{\eta},
\end{eqnarray}
that can be easily extrapolated as $n\to0$. 
The second equality follows from the normalization condition $\int p^{(\beta)}(\bm{x},\bm{z}|\bm{\eta})d\bm{x}d\bm{z}=1$.
Thus by obtaining tractable approximate densities for $p^{(\beta)}(\bm{x},\bm{z}|\bm{\eta})$ and $p^{(\beta)}(\bm{\eta})$ in (\ref{eq:de Finetti}), we can obtain an arbitrary degree of the moment without refitting\footnote{
    Of course, the replica symmetry may not hold for $n\notin\mathbb{N}$. In such cases, we have to include the effect of the replica symmetry breaking \cite{mezard1987spin}. However, we restrict ourselves to the replica symmetric case for simplicity.
}.

\subsection{Replica symmetric Gaussian expectation propagation in the replicated system}
To approximate the replicated system (\ref{eq:replicated system}), we will use the Gaussian diagonal EP of machine learning \cite{minka2001expectation, opper2005expectation} that is used to derive VAMP in \cite{rangan2019vector}. For $i=1,2,\dots,N$ and $\mu=1,2,\dots, M$, let $\tilde{\bm{x}}_i$ and $\tilde{\bm{z}}_\mu \in \mathbb{R}^n$ be $(x_{1,i}, x_{2,i},\dots, x_{n,i})^\top \in \mathbb{R}^n$ and $(z_{1,\mu}, z_{2,\mu}, \dots, z_{n, \mu})^\top \in \mathbb{R}^n$, respectively. The Gaussian diagonal EP recursively updates the following two approximate densities:
\begin{eqnarray}
\fl
    p_1^{(\beta)}(\{\bm{x}_{\a}\}_{\a=1}^n, \{\bm{z}_{\a}\}_{\a=1}^n) \propto \prod_{\mu=1}^M\mathbb{E}_{c_\mu}\left[
        \prod_{\a=1}^n p_{y|z}(y_\mu|z_{\a,\mu})^{\beta c_\mu}
    \right] \prod_{i=1}^N\mathbb{E}_{\gamma_i}\left[
        \prod_{\a=1}^n e^{-\beta \gamma_i |x_{\a, i}|}
    \right]
    \nonumber \\
    \times \underbrace{
        \prod_{i=1}^N e^{-\frac{1}{2}\tilde{\bm{x}}_i^\top \Lambda_{1x,i}^{(\beta)}\tilde{\bm{x}}_i + (\bm{h}_{1x,i}^{(\beta)})^\top \tilde{\bm{x}}_i} \prod_{\mu=1}^{M}e^{-\frac{1}{2}\tilde{\bm{z}}_{\mu}^\top \Lambda_{1z,\mu}^{(\beta)}\tilde{\bm{z}}_\mu + (\bm{h}_{1z,\mu}^{(\beta)})^\top \tilde{\bm{z}}_\mu}
    }_{\tilde{p}_1^{(\beta)}(\{\bm{x}_{\a}\}_{\a=1}^n, \{\bm{z}_{\a}\}_{\a=1}^n)},
    \label{eq:approximate factorized}
    \\
\fl
    p_2^{(\beta)}(\{\bm{x}_{\a}\}_{\a=1}^n, \{\bm{z}_{\a}\}_{\a=1}^n) \propto \prod_{\a=1}^n \delta(\bm{z}_{\a} - A\bm{x}_{\a})
    \nonumber \\
    \times \underbrace{
        \prod_{i=1}^N e^{-\frac{1}{2}\tilde{\bm{x}}_i^\top \Lambda_{2x,i}^{(\beta)}\tilde{\bm{x}}_i + (\bm{h}_{2x,i}^{(\beta)})^\top \tilde{\bm{x}}_i} \prod_{\mu=1}^{M}e^{-\frac{1}{2}\tilde{\bm{z}}_{\mu}^\top \Lambda_{2z,\mu}^{(\beta)}\tilde{\bm{z}}_\mu + (\bm{h}_{2z,\mu}^{(\beta)})^\top \tilde{\bm{z}}_\mu}
    }_{\tilde{p}_2^{(\beta)}(\{\bm{x}_{\a}\}_{\a=1}^n, \{\bm{z}_{\a}\}_{\a=1}^n)},
    \label{eq:approximate gaussian}
\end{eqnarray}
where $\Lambda_{1x,i}^{(\beta)}, \Lambda_{2x,i}^{(\beta)}, \Lambda_{1z, \mu}^{(\beta)}, \Lambda_{2z,\mu}^{(\beta)} \in \mathbb{R}^{n\times n}$ and $\bm{h}_{1x,i}^{(\beta)}, \bm{h}_{2x,i}^{(\beta)}, \bm{h}_{1z, \mu}^{(\beta)}, \bm{h}_{2z,\mu}^{(\beta)}\in\mathbb{R}^n$ are natural parameters of the Gaussians. The first approximation is a factorized distribution but contains the original non-Gaussian factors. The second approximation is a multivariate Gaussian distribution that replaces the non-Gaussian factors by the factorized Gaussians. Both of these distributions are tractable but ignore either the interactions or non-Gaussian factors. To include both the interactions and non-Gaussian factors, EP determines the natural parameters using the following moment-matching conditions:
\begin{eqnarray}
\fl
     \int x_{\a,i}p_1^{(\beta)}d^n\bm{x}d^n\bm{z} 
     &= 
     \int x_{\a,i}p_2^{(\beta)}d^n\bm{x}d^n\bm{z} 
     = \int x_{\a,i}\tilde{p}_{1}^{(\beta)}\tilde{p}_{2}^{(\beta)}d^n\bm{x}d^n\bm{z},
     \label{eq:first moment1}
     \\
\fl
     \int z_{\a,\mu}p_1^{(\beta)}d^n\bm{x}d^n\bm{z} 
     &= 
     \int z_{\a,\mu}p_2^{(\beta)}d^n\bm{x}d^n\bm{z} 
     = \int z_{\a,\mu}\tilde{p}_{1}^{(\beta)}\tilde{p}_{2}^{(\beta)}d^n\bm{x}d^n\bm{z}, 
     \label{eq:first moment2}
     \\
\fl
     \int x_{\a,i}x_{\b,i}p_1^{(\beta)}d^n\bm{x}d^n\bm{z} 
     &= 
     \int x_{\a,i}x_{\b,i}p_2^{(\beta)}d^n\bm{x}d^n\bm{z} 
     = \int x_{\a,i}x_{\b,i}\tilde{p}_{1}^{(\beta)}\tilde{p}_{2}^{(\beta)}d^n\bm{x}d^n\bm{z},
     \label{eq:second moment1}
     \\
\fl
     \int z_{\a,\mu}z_{\b,\mu}p_1^{(\beta)}d^n\bm{x}d^n\bm{z} 
     &= 
     \int z_{\a,\mu}z_{\b,\mu}p_2^{(\beta)}d^n\bm{x}d^n\bm{z} 
     = \int z_{\a,\mu}z_{\b,\mu}\tilde{p}_{1}^{(\beta)}\tilde{p}_{2}^{(\beta)}d^n\bm{x}d^n\bm{z},
     \label{eq:second moment2}
\end{eqnarray}
for any $i=1,2,\dots, N$, $\mu=1,2,\dots,M$, and $\a,\b=1,2,\dots,n$.  Schematically, the update rule of EP is depicted in algorithm \ref{algo:vanilla ep}. There, the density $\tilde{p}_1^{(\beta)}\tilde{p}_2^{(\beta)}$ is used to the moment-matching condition in lines \ref{line:vanilla ep moment_matching1}-\ref{line:vanilla ep moment_matching2} and \ref{line:vanilla ep moment_matching3}-\ref{line:vanilla ep moment_matching4}.

    \begin{algorithm}[p]
    \caption{Expectation propagation}
    \begin{algorithmic}[1]  \label{algo:vanilla ep}
    \REQUIRE{
        Approximate densities $p_1^{(\beta)}, p_2^{(\beta)}$ and the number of iterations $T_{\rm iter}$.
    }
    \STATE{Select initial $\Lambda_{1x,i}^{(\beta)}, \Lambda_{1z,\mu}^{(\beta)}, \bm{h}_{1x,i}^{(\beta)}$, and  $\bm{h}_{1z,\mu}^{(\beta)}$}
    \FOR{$t=1,2,\dots, T_{\rm iter}$}
        \STATE{// Factorized part (moment computation for $p_1^{(\beta)}$)}
        \vspace{1.5pt}
        \FOR{$i=1,2,\dots, N, \mu=1,2,\dots, M$ \label{line:vanilla ep factorized part first}}
            \STATE{
                $\hat{\bm{x}}_{1,i}^{(\beta)} = \int \tilde{\bm{x}}_i p_1^{(\beta)}d^n\bm{x}d^n\bm{z}$
            }
            \STATE{
                $\hat{\bm{z}}_{1,\mu}^{(\beta)} = \int \tilde{\bm{z}}_\mu p_1^{(\beta)}d\bm{x}d\bm{z}$
            } 
            \STATE{
                $V_{1x,i}^{(\beta)} = \int \tilde{\bm{x}}_i\tilde{\bm{x}}_i^\top p_1^{(\beta)}d^n\bm{x}d^n\bm{z} - (\hat{\bm{x}}_{1,i}^{(\beta)})(\hat{\bm{x}}_{1,i}^{(\beta)})^\top$
            } 
            \STATE{
                $V_{1z,\mu}^{(\beta)} = \int \tilde{\bm{z}}_\mu\tilde{\bm{z}}_\mu^\top p_1^{(\beta)}d^n\bm{x}d^n\bm{z} - (\hat{\bm{z}}_{1,\mu}^{(\beta)})(\hat{\bm{z}}_{1,\mu}^{(\beta)})^\top$
            } 
        \ENDFOR{\label{line:vanilla ep factorized part last}}
        \vspace{2.0pt}
        \STATE{// Moment-matching ($1\to2$) \label{line:vanilla ep moment_matching1}}
        \FOR{$i=1,2,\dots, N, \mu=1,2,\dots, M$}
            \STATE{
                update $\Lambda_{2x,i}^{(\beta)}, \Lambda_{2z,\mu}^{(\beta)}, \bm{h}_{2x,i}^{(\beta)}$ and $\bm{h}_{2z,\mu}^{(\beta)}$ so that the density $\tilde{p}_1^{(\beta)}\tilde{p}_2^{(\beta)}$ has the same moment with $p_1^{(\beta)}$ calculated in line \ref{line:vanilla ep factorized part first}-\ref{line:vanilla ep factorized part last}:
            }
            \STATE{\hspace{10pt}
                $\bm{h}_{2x,i}^{(\beta)} = (V_{1x,i}^{(\beta)})^{-1}\hat{\bm{x}}_{1,i}^{(\beta)} - \bm{h}_{1x,i}^{(\beta)}$
            }
            \STATE{\hspace{10pt}
                $\bm{h}_{2z,\mu}^{(\beta)} = (V_{1z,\mu}^{(\beta)})^{-1}\hat{\bm{z}}_{1,\mu}^{(\beta)} - \bm{h}_{1z,\mu}^{(\beta)}$
            }
            \STATE{\hspace{10pt}
                $\Lambda_{2x,i}^{(\beta)} = (V_{1x,i}^{(\beta)})^{-1} - \Lambda_{1x,i}^{(\beta)}$
            }
            \STATE{\hspace{10pt}
                $\Lambda_{2z,\mu}^{(\beta)} = (V_{1z,\mu}^{(\beta)})^{-1} - \Lambda_{1z,\mu}^{(\beta)}$
            }
        \ENDFOR{\label{line:vanilla ep moment_matching2}}
        \vspace{2.0pt}
        \STATE{// Gaussian part (moment computation for $p_{2}^{(\beta)}$)}
        \vspace{1.5pt}
        \FOR{$i=1,2,\dots, N, \mu=1,2,\dots, M$ \label{line:vanilla ep gaussian part first}}
            \STATE{
                $\hat{\bm{x}}_{2,i}^{(\beta)} = \int \tilde{\bm{x}}_i p_2^{(\beta)}d^n\bm{x}d^n\bm{z}$
            }
            \STATE{
                $\hat{\bm{z}}_{2,\mu}^{(\beta)} = \int \tilde{\bm{z}}_\mu p_2^{(\beta)}d\bm{x}d\bm{z}$
            } 
            \STATE{
                $V_{2x,i}^{(\beta)} = \int \tilde{\bm{x}}_i\tilde{\bm{x}}_i^\top p_2^{(\beta)}d^n\bm{x}d^n\bm{z} - (\hat{\bm{x}}_{2,i}^{(\beta)})(\hat{\bm{x}}_{2,i}^{(\beta)})^\top$
            } 
            \STATE{
                $V_{2z,\mu}^{(\beta)} = \int \tilde{\bm{z}}_\mu\tilde{\bm{z}}_\mu^\top p_2^{(\beta)}d^n\bm{x}d^n\bm{z} - (\hat{\bm{z}}_{2,\mu}^{(\beta)})(\hat{\bm{z}}_{2,\mu}^{(\beta)})^\top$
            } 
        \ENDFOR{\label{line:vanilla ep gaussian part last}}
        \vspace{2.0pt}
        \STATE{// Moment-matching ($2\to1$) \label{line:vanilla ep moment_matching3}}
        \vspace{1.5pt}
        \FOR{$i=1,2,\dots, N, \mu=1,2,\dots, M$}
            \STATE{
             update $\Lambda_{1x,i}^{(\beta)}, \Lambda_{1z,\mu}^{(\beta)}, \bm{h}_{1x,i}^{(\beta)}$ and $\bm{h}_{1z,\mu}^{(\beta)}$ so that the density $\tilde{p}_1^{(\beta)}\tilde{p}_2^{(\beta)}$ has the same moment with $p_2^{(\beta)}$ calculated in line \ref{line:vanilla ep gaussian part first}-\ref{line:vanilla ep gaussian part last}:
            } 
            \STATE{\hspace{10pt}
                $\bm{h}_{1x,i}^{(\beta)} = (V_{2x,i}^{(\beta)})^{-1}\hat{\bm{x}}_{2,i}^{(\beta)} - \bm{h}_{2x,i}^{(\beta)}$
            }
            \STATE{\hspace{10pt}
                $\bm{h}_{1z,\mu}^{(\beta)} = (V_{2z,\mu}^{(\beta)})^{-1}\hat{\bm{z}}_{2,\mu}^{(\beta)} - \bm{h}_{2z,\mu}^{(\beta)}$
            }
            \STATE{\hspace{10pt}
                $\Lambda_{1x,i}^{(\beta)} = (V_{2x,i}^{(\beta)})^{-1} - \Lambda_{2x,i}^{(\beta)}$
            }
            \STATE{\hspace{10pt}
                $\Lambda_{1z,\mu}^{(\beta)} = (V_{2z,\mu}^{(\beta)})^{-1} - \Lambda_{2z,\mu}^{(\beta)}$
            }
        \ENDFOR{}
    \ENDFOR{\label{line:vanilla ep moment_matching4}}
    \RETURN{$\Lambda_{1x,i}^{(\beta)}, \Lambda_{2x,i}^{(\beta)}, \Lambda_{1z, \mu}^{(\beta)}, \Lambda_{2z,\mu}^{(\beta)}$ and $\bm{h}_{1x,i}^{(\beta)}, \bm{h}_{2x,i}^{(\beta)}, \bm{h}_{1z, \mu}^{(\beta)}, \bm{h}_{2z,\mu}^{(\beta)}$.}
    
    \end{algorithmic}
    \end{algorithm}

The critical issue is to choose an appropriate form of the natural parameters in (\ref{eq:approximate factorized}) and (\ref{eq:approximate gaussian}). Based on the observations in subsection \ref{subsec:replica}, we impose the replica symmetry for these parameters:
\begin{eqnarray}
    \Lambda_{1x, i}^{(\beta)} &= \left(
            \begin{array}{ccc}
                 \beta \hat{Q}_{1x,i} - \beta^2 \hat{v}_{1x,i} & & -\beta^2 \hat{v}_{1x,i}  \\
                 & \ddots & \\
                 -\beta^2 \hat{v}_{1x,i} & & \beta \hat{Q}_{1x,i} - \beta^2 \hat{v}_{1x,i}
            \end{array}
        \right), 
    \label{eq:rs_l1x}
    \\
    \Lambda_{2x, i}^{(\beta)} &= \left(
        \begin{array}{ccc}
             \beta \hat{Q}_{2x,i} - \beta^2 \hat{v}_{2x,i} & & -\beta^2 \hat{v}_{2x,i}  \\
             & \ddots & \\
             -\beta^2 \hat{v}_{2x,i} & & \beta \hat{Q}_{2x,i} - \beta^2 \hat{v}_{2x,i}
        \end{array}
    \right), 
    \label{eq:rs_l2x}
    \\
    \Lambda_{1z, \mu}^{(\beta)} &= \left(
            \begin{array}{ccc}
                 \beta \hat{Q}_{1z,\mu} - \beta^2 \hat{v}_{1z,\mu} & & -\beta^2 \hat{v}_{1z,\mu}  \\
                 & \ddots & \\
                 -\beta^2 \hat{v}_{1z,\mu} & & \beta \hat{Q}_{1z,\mu} - \beta^2 \hat{v}_{1z,\mu}
            \end{array}
        \right), 
    \label{eq:rs_l1z}
    \\
    \Lambda_{2z, \mu}^{(\beta)} &= \left(
        \begin{array}{ccc}
             \beta \hat{Q}_{2z,\mu} - \beta^2 \hat{v}_{2z,\mu} & & -\beta^2 \hat{v}_{2z,\mu}  \\
             & \ddots & \\
             -\beta^2 \hat{v}_{2z,\mu} & & \beta \hat{Q}_{2z, \mu} - \beta^2 \hat{v}_{2z,\mu}
        \end{array}
    \right), 
    \label{eq:rs_l2z}
    \\
    \bm{h}_{1x,i}^{(\beta)} &= \beta h_{1x,i}\bm{1}_N,
    \label{eq:rs_h1x}
    \\
    \bm{h}_{2x,i}^{(\beta)} &= \beta h_{2x,i}\bm{1}_N,
    \label{eq:rs_h2x}
    \\
    \bm{h}_{1z,\mu}^{(\beta)} &= \beta h_{1z,\mu}\bm{1}_M,
    \label{eq:rs_h1z}
    \\
    \bm{h}_{2z,\mu}^{(\beta)} &= \beta h_{2z,\mu}\bm{1}_M.
    \label{eq:rs_h2z}
\end{eqnarray}
With these parameterizations, we use $\hat{\bm{Q}}_{1x}=(\hat{Q}_{1x,1}, \hat{Q}_{1x,2}, \dots, \hat{Q}_{1x,N})^\top$ for the vector notation. $\hat{\bm{Q}}_{2x}, \hat{\bm{Q}}_{1z}, \hat{\bm{Q}}_{2z}, \hat{\bm{v}}_{1x}, \hat{\bm{v}}_{2x}, \hat{\bm{v}}_{1z}, \hat{\bm{v}}_{2z}, \bm{h}_{1x}, \bm{h}_{2x}, \bm{h}_{1z}$, and $\bm{h}_{2z}$ are defined similarly. These parameterizations allow the extrapolation $n\to0$ as follows.

For $\eta_{x,i},\eta_{z,\mu}\in\mathbb{R}$, let $\phi_{x,i}^{(\beta)}$ and $\phi_{z,\mu}^{(\beta)}$ be 
\begin{eqnarray}
\fl
    \phi_{x,i}^{(\beta)} &= \frac{1}{\beta}\log \int
        \exp\left(
            -\beta\frac{\hat{Q}_{1x,i}}{2}x^2 
            + \beta(h_{1x,i} + \sqrt{\hat{v}_{1x,i}}\eta_{x,i})x
            -\beta\gamma_i|x|
        \right)
    dx,
    \\
\fl
    \phi_{z,\mu}^{(\beta)} &= \frac{1}{\beta}\log\int
        \exp\left(
            -\beta\frac{\hat{Q}_{1z,\mu}}{2}z^2
            + \beta(h_{1z,\mu} + \sqrt{ \hat{v}_{1z,\mu} } \eta_{z,\mu})z
            +\beta c_\mu \log p_{y|z}(y_\mu|z)
        \right)
    dz.
\end{eqnarray}
We also denote by $Dx=e^{-x^2/2}/\sqrt{2\pi}$ the standard Gaussian measure, and by $\diag{\bm{x}}$ a diagonal matrix with $[\diag{\bm{x}}]_{ii}=x_i$. The use of the replica symmetric parameterizations (\ref{eq:rs_l1x})-(\ref{eq:rs_h2z}) yields the following expressions for the moments and the moment-matching conditions that are used in line \ref{line:vanilla ep moment_matching1}-\ref{line:vanilla ep moment_matching2} and \ref{line:vanilla ep moment_matching3}-\ref{line:vanilla ep moment_matching4} in algorithm \ref{algo:vanilla ep}. First, for the approximate density $p_1^{(\beta)}$, we obtain
\begin{eqnarray}
    \int x_{\a,i}p_1^{(\beta)}d^n\bm{x}d^n\bm{z} = \hat{x}_{1,i}, 
    \\
    \int x_{\a,i}x_{\b,i} p_1^{(\beta)}d^n\bm{x}d^n\bm{z} = v_{1x,i} +  \hat{x}_{1,i}^2, \, \a\neq\b,
    \\
    \int x_{\a,i}^2 p_1^{(\beta)}d^n\bm{x}d^n\bm{z} = \frac{\chi_{1x,i}}{\beta} + v_{1x,i} + \hat{x}_{1,i}^2, 
    \\
    \int z_{\a,\mu}p_1^{(\beta)}d^n\bm{x}d^n\bm{z} = \hat{z}_{1,\mu}, 
    \\
    \int z_{\a,\mu}z_{\b,\mu} p_1^{(\beta)}d^n\bm{x}d^n\bm{z} = v_{1z,\mu} +  \hat{z}_{1,\mu}^2, \, \a\neq\b,
    \\
    \int z_{\a,\mu}^2 p_1^{(\beta)}d^n\bm{x}d^n\bm{z} = \frac{\chi_{1z,\mu}}{\beta} + v_{1z,\mu} + \hat{z}_{1,\mu}^2, 
\end{eqnarray}
where 
\begin{eqnarray}
    \hat{x}_{1,i} 
    = 
    \frac{
        \mathbb{E}_{\gamma_i}\left[
            \int 
                \frac{
                    \partial \phi_{x,i}^{(\beta)}
                }{
                    \partial h_{1x,i}
                }
            e^{\beta n\phi_{x,i}^{(\beta)}}D\eta_{x,i}
        \right]
    }{
        \mathbb{E}_{\gamma_i}\left[
            \int e^{\beta n\phi_{x,i}^{(\beta)}}D\eta_{x,i}
        \right]
    }, 
    \\
    \chi_{1x,i} = \frac{
        \mathbb{E}_{\gamma_i}\left[
            \int 
                \frac{
                    \partial^2 \phi_{x,i}^{(\beta)}
                }{
                    \partial h_{1x,i}^2
                }
            e^{\beta n\phi_{x,i}^{(\beta)}}D\eta_{x,i}
        \right]
    }{
        \mathbb{E}_{\gamma_i}\left[
            \int e^{\beta n\phi_{x,i}^{(\beta)}}D\eta_{x,i}
        \right]
    }, 
    \\
\fl
    v_{1x,i} = 
    \frac{
        \mathbb{E}_{\gamma_i}\left[
            \int 
                \left(
                    \frac{
                        \partial \phi_{x,i}^{(\beta)}
                    }{
                        \partial h_{1x,i}
                    }
                \right)^2
            e^{\beta n\phi_{x,i}^{(\beta)}}D\eta_{x,i}
        \right]
    }{
        \mathbb{E}_{\gamma_i}\left[
            \int e^{\beta n\phi_{x,i}^{(\beta)}}D\eta_{x,i}
        \right]
    }
    -
    \left(
        \frac{
            \mathbb{E}_{\gamma_i}\left[
                \int 
                    \frac{
                        \partial \phi_{x,i}^{(\beta)}
                    }{
                        \partial h_{1x,i}
                    }
                e^{\beta n\phi_{x,i}^{(\beta)}}D\eta_{x,i}
            \right]
        }{
            \mathbb{E}_{\gamma_i}\left[
                \int e^{\beta n\phi_{x,i}^{(\beta)}}D\eta_{x,i}
            \right]
        }
    \right)^2
    , 
    \\
    \hat{z}_{1,\mu}
    = 
    \frac{
        \mathbb{E}_{c_{\mu}}\left[
            \int 
                \frac{
                    \partial \phi_{z,\mu}^{(\beta)}
                }{
                    \partial h_{1z,\mu}
                }
            e^{\beta n\phi_{z,\mu}^{(\beta)}}D\eta_{z,\mu}
        \right]
    }{
        \mathbb{E}_{c_{\mu}}\left[
            \int e^{\beta n\phi_{z,\mu}^{(\beta)}}D\eta_{z,\mu}
        \right]
    }, 
    \\
    \chi_{1z,\mu} = \frac{
        \mathbb{E}_{c_\mu}\left[
            \int 
                \frac{
                    \partial^2 \phi_{z,\mu}^{(\beta)}
                }{
                    \partial h_{1z,\mu}^2
                }
            e^{\beta n\phi_{z,\mu}^{(\beta)}}D\eta_{z,\mu}
        \right]
    }{
        \mathbb{E}_{c_\mu}\left[
            \int e^{\beta n\phi_{z,\mu}^{(\beta)}}D\eta_{z,\mu}
        \right]
    }, 
    \\
\fl
    v_{1z,\mu} = 
    \frac{
        \mathbb{E}_{c_\mu}\left[
            \int 
                \left(
                    \frac{
                        \partial \phi_{z,\mu}^{(\beta)}
                    }{
                        \partial h_{1z,\mu}
                    }
                \right)^2
            e^{\beta n\phi_{z,\mu}^{(\beta)}}D\eta_{z,\mu}
        \right]
    }{
        \mathbb{E}_{c_\mu}\left[
            \int e^{\beta n\phi_{z,\mu}^{(\beta)}}D\eta_{z,\mu}
        \right]
    }
    -
    \left(
        \frac{
            \mathbb{E}_{c_\mu}\left[
                \int 
                    \frac{
                        \partial \phi_{z,\mu}^{(\beta)}
                    }{
                        \partial h_{1z,\mu}
                    }
                e^{\beta n\phi_{z,\mu}^{(\beta)}}D\eta_{z,\mu}
            \right]
        }{
            \mathbb{E}_{c_\mu}\left[
                \int e^{\beta n\phi_{z,\mu}^{(\beta)}}D\eta_{z,\mu}
            \right]
        }
    \right)^2
    ,
\end{eqnarray}
Next, for the approximate density $p_2^{(\beta)}$, we obtain
\begin{eqnarray}
    \int x_{\a,i}p_2^{(\beta)}d^n\bm{x}d^n\bm{z} = \hat{x}_{2,i}, 
    \\
    \int x_{\a,i}x_{\b,i} p_2^{(\beta)}d^n\bm{x}d^n\bm{z} = v_{2x,i} +  \hat{x}_{2,i}^2, \, \a\neq\b,
    \\
    \int x_{\a,i}^2 p_2^{(\beta)}d^n\bm{x}d^n\bm{z} = \frac{\chi_{2x,i}}{\beta} + v_{2x,i} + \hat{x}_{2,i}^2, 
    \\
    \int z_{\a,\mu}p_2^{(\beta)}d^n\bm{x}d^n\bm{z} = \hat{z}_{2,\mu}, 
    \\
    \int z_{\a,\mu}z_{\b,\mu} p_2^{(\beta)}d^n\bm{x}d^n\bm{z} = v_{2z,\mu} +  \hat{z}_{2,\mu}^2, \, \a\neq\b,
    \\
    \int z_{\a,\mu}^2 p_2^{(\beta)}d^n\bm{x}d^n\bm{z} = \frac{\chi_{2z,\mu}}{\beta} + v_{2z,\mu} + \hat{z}_{2,\mu}^2, 
\end{eqnarray}
where 
\begin{eqnarray}
    \hat{\bm{x}}_{2}
    = 
    \left(
        \diag{\hat{\bm{Q}}_{2x}} + A^\top \diag{\hat{\bm{Q}}_{2z}}A
    \right)^{-1}
    \left(
        \bm{h}_{2x} + A^\top \bm{h}_{2z}
    \right)
    , 
    \\
    \chi_{2x,i}
    =
    \left[
        \left(
            \diag{\hat{\bm{Q}}_{2x}} + A^\top \diag{\hat{\bm{Q}}_{2z}}A
        \right)^{-1}
    \right]_{ii}
    ,
    \\
\fl
    v_{2x,i}
    =
    \left[
        \left(
            \diag{\hat{\bm{Q}}_{2x}} + A^\top \diag{\hat{\bm{Q}}_{2z}}A
        \right)^{-1}
        \left(
            \diag{\hat{\bm{v}}_{2x}} + A^\top \diag{ \hat{\bm{v}}_{2z}} A
        \right)
    \right.
    \nonumber \\
    \left.
        \times 
        \left(
            \diag{\hat{\bm{Q}}_{2x}} + A^\top \diag{\hat{\bm{Q}}_{2z}}A
        \right)^{-1}
    \right]_{ii},
    \\
    \hat{\bm{z}}_2
    =
    A^\top\hat{\bm{x}_2}
    ,
    \\
    \chi_{2z,\mu}
    =
    \left[
        A
        \left(
            \diag{\hat{\bm{Q}}_{2x}} + A^\top \diag{\hat{\bm{Q}}_{2z}}A
        \right)^{-1}
        A^\top
    \right]_{\mu\mu}
    ,
    \\
\fl
    v_{2z,\mu}
    =
    \left[
        A
        \left(
            \diag{\hat{\bm{Q}}_{2x}} + A^\top \diag{\hat{\bm{Q}}_{2z}}A
        \right)^{-1}
        \left(
            \diag{\hat{\bm{v}}_{2x}} + A^\top \diag{ \hat{\bm{v}}_{2z}} A
        \right)
    \right.
    \nonumber \\
    \left.
        \times  
        \left(
            \diag{\hat{\bm{Q}}_{2x}} + A^\top \diag{\hat{\bm{Q}}_{2z}}A
        \right)^{-1}
        A^\top
    \right]_{\mu\mu}.
\end{eqnarray}
Finally, the moment-matching conditions are written as 
\begin{eqnarray}
    h_{2x,i} &= \frac{\hat{x}_{1,i}}{\chi_{1x,i}} - h_{1x,i} + \mathcal{O}(n)
    , \quad
    h_{1x,i} &= \frac{\hat{x}_{2,i}}{\chi_{2x,i}} - h_{2x,i} + \mathcal{O}(n)
    ,
    \\
    \hat{Q}_{2x,i} &= \frac{1}{\chi_{1x,i}} - \hat{Q}_{1x,i} + \mathcal{O}(n)
    , \quad
    \hat{Q}_{1x,i} &= \frac{1}{\chi_{2x,i}} - \hat{Q}_{2x,i} + \mathcal{O}(n)
    ,
    \\
    \hat{v}_{2x,i} &= \frac{v_{1x,i}}{\chi_{1x,i}^2} - \hat{v}_{1x,i} + \mathcal{O}(n)
    , \quad
    \hat{v}_{1x,i} &= \frac{v_{2x,i}}{\chi_{2x,i}^2} - \hat{v}_{2x,i} + \mathcal{O}(n)
    ,
    \\
    h_{2z,\mu} &= \frac{\hat{z}_{1,\mu}}{\chi_{1z,\mu}} - h_{1z,\mu} + \mathcal{O}(n)
    , \quad
    h_{1z,\mu} &= \frac{\hat{z}_{2,\mu}}{\chi_{2z,\mu}} - h_{2z,\mu} + \mathcal{O}(n)
    ,
    \\
    \hat{Q}_{2z,\mu} &= \frac{1}{\chi_{1z,\mu}} - \hat{Q}_{1z,\mu} + \mathcal{O}(n)
    , \quad
    \hat{Q}_{1z,\mu} &= \frac{1}{\chi_{2z,\mu}} - \hat{Q}_{2z,\mu} + \mathcal{O}(n)
    ,
    \\
    \hat{v}_{2z,\mu} &= \frac{v_{1z,\mu}}{\chi_{1z,\mu}^2} - \hat{v}_{1z,\mu} + \mathcal{O}(n)
    , \quad
    \hat{v}_{1z,\mu} &= \frac{v_{2z,\mu}}{\chi_{2z,\mu}^2} - \hat{v}_{2z,\mu} + \mathcal{O}(n)
    ,
\end{eqnarray}
In all of the above expressions, the indices $i$ and $\mu$ run as $i=1,2,\dots, N$ and $\mu=1,2,\dots, M$, respectively. $\bm{\chi}_x$ and $\bm{\chi}_{z}$ are termed susceptibility. $\bm{v}_{x}$ and $\bm{v}_z$ are termed variance. Clearly, these equations can be easily extrapolated as $n\to0$.

    \begin{algorithm}[t]
    \caption{rVAMP}
    \begin{algorithmic}[1]  \label{algo:rVAMP}
    \REQUIRE{
        Denoising functions $g_{1x}, g_{1z}$ from (\ref{eq:denoiser1x}) and (\ref{eq:denoiser1z}), the features $A\in\mathbb{R}^{M\times N}$, the response variable $\bm{y}\in\mathcal{Y}^M$, the convergence criterion $\epsilon_{\rm tol}$, the maximum number of iterations $T_{\rm iter}$.
    }
    \STATE{Select initial $\bm{h}_{1x}^{(1)}\in\mathbb{R}^N, \bm{h}_{1z}^{(1)}\in\mathbb{R}^M$, $\hat{\bm{Q}}_{1x}^{(1)}, \hat{\bm{v}}_{1x}^{(1)}\in[0,\infty)^N$, and $\hat{\bm{Q}}_{1z}^{(1)}, \hat{\bm{v}}_{1z}^{(1)}\in[0,\infty)^M$.}
    \FOR{$t=1,2,\dots, T_{\rm iter}$}
        \STATE{// Factorized part}
        \vspace{2.0pt}
        \STATE{$
                \hat{\bm{x}}_{1}^{(t)} = \mathbb{E}_{\bm{\gamma}}[\int \bm{g}_{1x}(\bm{h}_{1x}^{(t)}, \hat{\bm{Q}}_{1x}^{(t)}, \hat{\bm{v}}_{1x}^{(t)}; \bm{\gamma}, \bm{\eta}_{x})D\bm{\eta}_{x}]
        $
        \label{line:factorized x1}
        }
        \vspace{2.0pt}
        \STATE{$
                \bm{\chi}_{1x}^{(t)} = \mathbb{E}_{\bm{\gamma}}[\int \bm{g}_{1x}^\prime(\bm{h}_{1x}^{(t)}, \hat{\bm{Q}}_{1x}^{(t)}, \hat{\bm{v}}_{1x}^{(t)}; \bm{\gamma}, \bm{\eta}_{x})D\bm{\eta}_{x}]
        $
        \label{line:factorized x2}
        }
        \vspace{2.0pt}
        \STATE{$
                \bm{v}_{1x}^{(t)} = \mathbb{E}_{\bm{\gamma}}[\int \bm{g}_{1x}^2(\bm{h}_{1x}^{(t)}, \hat{\bm{Q}}_{1x}^{(t)}, \hat{\bm{v}}_{1x}^{(t)}; \bm{\gamma}, \bm{\eta}_{x})D\bm{\eta}_{x}]
                - (\hat{\bm{x}}_{1}^{(t)})^2
        $
        \label{line:factorized x3}
        }
        \STATE{$
                \hat{\bm{z}}_{1}^{(t)} = \mathbb{E}_{\bm{c}}[\int \bm{g}_{1z}(\bm{h}_{1z}^{(t)}, \hat{\bm{Q}}_{1z}^{(t)}, \hat{\bm{v}}_{1z}^{(t)}; \bm{c}, \bm{\eta}_{z}, \bm{y})D\bm{\eta}_{z}]
        $
        \label{line:factorized z1}
        }
        \vspace{2.0pt}
        \STATE{$
                \bm{\chi}_{1z}^{(t)} = \mathbb{E}_{\bm{c}}[\int \bm{g}_{1z}^\prime(\bm{h}_{1z}^{(t)}, \hat{\bm{Q}}_{1z}^{(t)}, \hat{\bm{v}}_{1z}^{(t)}; \bm{c}, \bm{\eta}_{z}, \bm{y})D\bm{\eta}_{z}]
        $
        \label{line:factorized z2}
        }
        \vspace{2.0pt}
        \STATE{$
                \bm{v}_{1z}^{(t)} = \mathbb{E}_{\bm{c}}[\int \bm{g}_{1z}^2(\bm{h}_{1z}^{(t)}, \hat{\bm{Q}}_{1z}^{(t)}, \hat{\bm{v}}_{1z}^{(t)}; \bm{c}, \bm{\eta}_{z}, \bm{y})D\bm{\eta}_{z}]
                - (\hat{\bm{z}}_{1}^{(t)})^2
        $
        \label{line:factorized z3}
        }
        \vspace{1.5pt}
        \STATE{// Moment-matching ($1\to2$)}
        \vspace{2.0pt}
        \STATE{$
            \bm{h}_{2x}^{(t)} = \hat{\bm{x}}_{1}^{(t)}/\bm{\chi}_{1x}^{(t)} - \bm{h}_{1x}^{(t)}
            ,\quad
            \hat{\bm{Q}}_{2x}^{(t)} = \left(\bm{\chi}_{1x}^{(t)}\right)^{-1} - \hat{\bm{Q}}_{1x}^{(t)}
            ,\quad
            \hat{\bm{v}}_{2x}^{(t)} = \bm{v}_{1x}^{(t)} / \left( \bm{\chi}_{1x}^{(t)} \right)^2 - \hat{\bm{v}}_{1x}^{(t)}
        $}
        \STATE{$
                \bm{h}_{2z}^{(t)} = \hat{\bm{z}}_{1}^{(t)}/\bm{\chi}_{1z}^{(t)} - \bm{h}_{1z}^{(t)}
                ,\quad
                \hat{\bm{Q}}_{2z}^{(t)} = \left(\bm{\chi}_{1z}^{(t)}\right)^{-1} - \hat{\bm{Q}}_{1z}^{(t)}
                ,\quad
                \hat{\bm{v}}_{2z}^{(t)} = \bm{v}_{1z}^{(t)} / \left( \bm{\chi}_{1z}^{(t)} \right)^2 - \hat{\bm{v}}_{1z}^{(t)}
            $}
        \vspace{2.0pt}
        \STATE{// Gaussian part}
        \STATE{$
            X = \left(
                    \diag{\hat{\bm{Q}}_{2x}^{(t)}} + A^\top \diag{\hat{\bm{Q}}_{2z}}A
                \right)^{-1}
        $
        \label{line:inverse}
        }
        \STATE{$
            \hat{\bm{x}}_2^{(t)} = X(\bm{h}_{2x}^{(t)}+ A^\top \bm{h}_{2z}^{(t)})
            , \quad
            \hat{\bm{z}}_2^{(t)} = A\hat{\bm{x}}_2^{(t)}
        $}
        \STATE{$
            \bm{\chi}_{2x}^{(t)} = \mathop{\rm diag}[X]
            , \quad
            \bm{\chi}_{2z}^{(t)} = \mathop{\rm diag}[AXA^\top]
        $}
        \vspace{1.5pt}
        \STATE{$
            \bm{v}_{2x}^{(t)} = \mathop{\rm diag}\left[
                    X\left(
                        \diag{\hat{\bm{v}}_{2x}^{(t)}} + A^\top \diag{\hat{\bm{v}}_{2z}^{(t)}}A
                    \right)X
                \right]
        $}
        \vspace{1.5pt}
        \STATE{$
            \bm{v}_{2z}^{(t)} = \mathop{\rm diag}\left[
                AX\left(
                    \diag{\hat{\bm{v}}_{2x}^{(t)}} + A^\top \diag{\hat{\bm{v}}_{2z}^{(t)}}A
                \right)XA^\top
            \right]
        $}
        \STATE{// Moment-matching ($2\to1$)}
        \STATE{$
            \bm{h}_{1x}^{(t+1)} = \hat{\bm{x}}_{2}^{(t)}/\bm{\chi}_{2x}^{(t)} - \bm{h}_{2x}^{(t)}
            ,\quad
            \hat{\bm{Q}}_{1x}^{(t+1)} = \left(\bm{\chi}_{2x}^{(t)}\right)^{-1} - \hat{\bm{Q}}_{2x}^{(t)}
            ,\quad
            \hat{\bm{v}}_{1x}^{(t+1)} = \bm{v}_{2x}^{(t)} / \left( \bm{\chi}_{2x}^{(t)} \right)^2 - \hat{\bm{v}}_{2x}^{(t)}
        $
        \label{line:rVAMP moment-matching 2to1 1}
        }
        \STATE{$
                \bm{h}_{1z}^{(t+1)} = \hat{\bm{z}}_{2}^{(t)}/\bm{\chi}_{2z}^{(t)} - \bm{h}_{2z}^{(t)}
                ,\quad
                \hat{\bm{Q}}_{1z}^{(t+1)} = \left(\bm{\chi}_{2z}^{(t)}\right)^{-1} - \hat{\bm{Q}}_{2z}^{(t)}
                ,\quad
                \hat{\bm{v}}_{1z}^{(t+1)} = \bm{v}_{2z}^{(t)} / \left( \bm{\chi}_{2z}^{(t)} \right)^2 - \hat{\bm{v}}_{2z}^{(t)}
            $
        \label{line:rVAMP moment-matching 2to1 2}
        }
        \IF{$\max\{\|\hat{\bm{x}}_1^{(t)} - \hat{\bm{x}}_2^{(t)}\|_2^2/N, \|\hat{\bm{z}}_1^{(t)} - \hat{\bm{z}}_2^{(t)}\|_2^2/M\} < \epsilon_{\rm tol}$}
            \STATE{$t\leftarrow T_{\rm iter}$}
            \STATE{break}
        \ENDIF
    \ENDFOR
    \RETURN{$\bm{h}_{1x}^{(T_{\rm iter})}, \hat{\bm{Q}}_{1x}^{(T_{\rm iter})},\hat{\bm{v}}_{1x}^{(T_{\rm iter})}$}
    \end{algorithmic}
    \end{algorithm}
Inserting the limiting form of these quantities at $n\to0, \beta\to\infty$ into the algorithm \ref{algo:vanilla ep}, we obtain rVAMP in algorithm \ref{algo:rVAMP}. There, $\bm{g}_{1x}, \bm{g}_{1z}, \bm{g}_{1x}^\prime$ and $\bm{g}_{1z}^\prime$ are denoising functions and their derivatives. These are defined as follows:
\begin{eqnarray}
     \bm{g}_{1x}(\bm{h}_{1x}, \hat{\bm{Q}}_{1x}, \hat{\bm{v}}_{1x}; \bm{\gamma}, \bm{\eta}_{x})
     = [g_{1x}(h_{1x,i}, \hat{Q}_{1x,i}, \hat{v}_{1x,i}; \gamma_i, \eta_{x,i})]_{1\le i\le N},
     \label{eq:denoiser1x}
     \\
     \bm{g}_{1x}^\prime(\bm{h}_{1x}, \hat{\bm{Q}}_{1x}, \hat{\bm{v}}_{1x}; \bm{\gamma}, \bm{\eta}_{x})
     = [g_{1x}^\prime(h_{1x,i}, \hat{Q}_{1x,i}, \hat{v}_{1x,i}; \gamma_i, \eta_{x,i})]_{1\le i\le N},
     \\
    \bm{g}_{1z}(\bm{h}_{1z}, \hat{\bm{Q}}_{1z}, \hat{\bm{v}}_{1z}; \bm{c}, \bm{\eta}_{z}, \bm{y})
    = [g_{1z}(h_{1z,\mu}, \hat{Q}_{1z,\mu}, \hat{v}_{1z,\mu}; c_\mu, \eta_{z,\mu}, y_\mu)]_{1\le \mu\le M},
    \label{eq:denoiser1z}
    \\
    \bm{g}_{1z}^\prime(\bm{h}_{1z}, \hat{\bm{Q}}_{1z}, \hat{\bm{v}}_{1z}; \bm{c}, \bm{\eta}_{z}, \bm{y})
    = [g_{1z}^\prime(h_{1z,\mu}, \hat{Q}_{1z,\mu}, \hat{v}_{1z,\mu}; c_\mu, \eta_{z,\mu}, y_\mu)]_{1\le \mu\le M},
\end{eqnarray}
where
\begin{eqnarray}
\fl
    g_{1x}(h_{1x,i}, \hat{Q}_{1x,i}, \hat{v}_{1x,i}; \gamma_i, \eta_{x,i})
    = \frac{h_{1x,i} + \sqrt{\hat{v}}_{1x,i}\eta_{x,i} -\gamma_i\mathop{\rm sign}(h_{1x,i} + \sqrt{\hat{v}_{1x,i}}\eta_{x,i} )}{\hat{Q}_{1x,i}}
    \nonumber \\
    \times \1\left(
        \left|
            h_{1x,i} + \sqrt{\hat{v}_{1x,i}}\eta_{x,i}
        \right| 
        > \gamma_i
    \right),
    \\
\fl
    g_{1x}^\prime(h_{1x,i}, \hat{Q}_{1x,i}, \hat{v}_{1x,i}; \gamma_i, \eta_{x,i})
    = \frac{1}{\hat{Q}_{1x,i}}
    \1\left(
        \left|
            h_{1x,i} + \sqrt{\hat{v}_{1x,i}}\eta_{x,i}
        \right| 
        > \gamma_i
    \right)
    , \\
\fl
    g_{1z}(h_{1z,\mu}, \hat{Q}_{1z,\mu}, \hat{v}_{1z,\mu}; c_\mu, \eta_{z,\mu}, y_\mu) = \arg\max_{z\in\mathbb{R}}\left[
        -\frac{\hat{Q}_{1z,\mu}}{2}z^2 + (h_{1z,\mu} 
    \right.
    \nonumber \\
    \left.
        + \sqrt{\hat{v}_{1z,\mu}}\eta_{z,\mu})z + c_\mu \log p_{y|z}(y_\mu|z)
    \right],
    \\
\fl
    g_{1z}^\prime(h_{1z,\mu}, \hat{Q}_{1z,\mu}, \hat{v}_{1z,\mu}; c_\mu, \eta_{z,\mu}, y_\mu) = \frac{\partial g_{1z}(h_{1z,\mu}, \hat{Q}_{1z,\mu}, \hat{v}_{1z,\mu}; c_\mu, \eta_{z,\mu}, y_\mu)}{\partial h_{1z,\mu}}.
\end{eqnarray}
If the likelihood $p_{y|z}$ is differentiable with respect to $z$, $g_{1z}^\prime$ can be written as 
\begin{equation}
\fl
    g_{1z}^\prime(h_{1z,\mu}, \hat{Q}_{1z,\mu}, \hat{v}_{1z,\mu}; c_\mu, \eta_{z,\mu}, y_\mu) = \left[
        \hat{Q}_{1z,\mu} - c_\mu \left.
            \frac{\partial^2 \log p_{y|z}(y_\mu|z)}{\partial z^2}
        \right|_{z=g_{1z}}
    \right]^{-1}.
    \label{eq:chi1z_susceptibility}
\end{equation}
Because the averages with respect to $\bm{c}$ and $\bm{\gamma}$ are incorporated in line \ref{line:factorized x1}-\ref{line:factorized z3} of the algorithm \ref{algo:rVAMP} as the averages with respect to one-dimensional random variables, rVAMP does not require refitting.

Although the two approximate densities have the same first and second moments at a fixed point, these two densities have different characteristics. For higher-order marginal moments, we expect that $p_1^{(\beta)}$ is more precise than $p_2^{(\beta)}$ because it accurately includes the non-Gaussian factors. Similarly, $p_2^{(\beta)}$ is argued to have more accurate off-diagonal moments because it includes the interaction term correctly \cite{opper2004variational, opper2005expectation}. Thus, these two distributions should be used depending on the objective. Because we are interested in the distribution of the marginal moment (\ref{eq:marginal_dist}), here we use $p_1^{(\beta)}$ to compute $\Pi_i(\gamma_0)$. 

\subsection{Calculation of the selection probability}
Using the expression
\begin{eqnarray}
\fl
    p_1^{(\beta)}(\{\bm{x}_\a\}, \{\bm{z}_\a\})
    &\propto \prod_{i=1}^N \mathbb{E}_{\gamma_i}\left[
        \int 
            \prod_{\a=1}^n
            e^{
                -\frac{\beta\hat{Q}_{1x,i}}{2}x_{\a,i}^2
                +\beta(
                    h_{1x,i} + \sqrt{\hat{v}_{1x,i}}\eta_{x,i}
                )x_{\a,i}
                -\beta\gamma_i|x_{\a,i}|
            }
        D\eta_{x,i}
    \right]
    \nonumber \\
\fl 
    & \hspace{-20pt} \times\prod_{\mu=1}^M \mathbb{E}_{c_\mu}\left[
        \int 
            \prod_{\a=1}^n
            e^{
                -\frac{\beta \hat{Q}_{1z,\mu}}{2}z_{\a,\mu}^2
                + \beta (
                    h_{1z,\mu} + \sqrt{\hat{v}_{1z,\mu}}z_{\a, \mu}
                )
            }
            p_{y|z}(y_\mu|z_{\a,\mu})^{\beta c_\mu}
        D\eta_{z,\mu}
    \right],
\end{eqnarray}
we obtain the following form of the $r$-th moment:
\begin{equation}
    \mathbb{E}_{\bm{c},\bm{\gamma}}\left[\hat{x}_i^r\right]
    = \mathbb{E}_{\gamma_i}\left[
        \int
            g_{1x}(h_{1x,i},\hat{Q}_{1x,i},\hat{v}_{1x,i};\gamma_i, \eta_{x,i})^r
        D\eta_{x,i}
    \right].
    \label{eq:r-th-moment-rvamp}
\end{equation}
To understand the meaning of $\eta_{x,i}$, suppose that we omit to take the expectations of $(\bm{c}, \bm{\gamma})$ in lines \ref{line:factorized x1}-\ref{line:factorized z3} of algorithm 2 and to run rVAMP for a fixed set of $(\bm{c}, \bm{\gamma})$. Then, one can show that $v_{1x,i}=v_{2x,i}=\hat{v}_{1x,i}=\hat{v}_{2x,i}=0$ and $v_{1z,\mu}=v_{2z,\mu}=\hat{v}_{1z,\mu}=\hat{v}_{2z,\mu}=0$ yield the fixed point condition for these variables, and the rest part of the algorithm exactly coincides with the VAMP algorithm for LASSO without a resampling \cite{rangan2019vector}. Thus, we expect that $\sqrt{\hat{v}_{1x,i}}\eta_{x,i}$ behave as random variables that approximately reflect the effect of taking average of $\bm{c}$. This consideration and the expression of the $r$-th moment in (\ref{eq:r-th-moment-rvamp}) yield the following form of the distribution function $p(m_i)$:
\begin{equation}
    p(m_i) \simeq \mathbb{E}_{\gamma_i}\left[
        \int 
            \1\left(
                m_{i}
                -
                g_{1x}(h_{1x, i}, \hat{Q}_{1x,i}, \hat{v}_{1x,i};\gamma_i, \eta_{x,i})
            \right)
        D\eta_{x,i}
    \right].
\end{equation}
Because $g_{1x}(h_{1x, i}, \hat{Q}_{1x,i}, \hat{v}_{1x,i};\gamma_i, \eta_{x,i})$ is non-zero iff $\1(|h_{1x,i} + \sqrt{\hat{v}_{1x,i}}\eta_{x,i}|>\gamma_i)$ is satisfied, rVAMP yields the following expression for the selection probability $\Pi_i$:
\begin{equation}
    \Pi_i(\gamma_0) \simeq \mathbb{E}_{\gamma_i}
    \left[
        \int 
            \1\left(
                \left|
                    h_{1x,i} + \sqrt{\hat{v}_{1x,i}}\eta_{x,i}
                \right| 
                > \gamma_i
            \right)
        D\eta_{x,i}
    \right],
\end{equation}
which is easy to calculate.

\subsection{Implementation details}
\label{subsection:implementation details}

For practical implementation, we find that it is helpful to make several small modifications to rVAMP of the algorithm \ref{algo:rVAMP}. In this subsection, we discuss these minor modifications.

First we address the computational complexity regarding the matrix inversion. Although rVAMP requires the matrix inversion in line \ref{line:inverse}, this computational cost is reduced to $\mathcal{O}(M^3)$ from $\mathcal{O}(N^3)$ using the Woodbury identity \cite{golub1996matrix}:
\begin{eqnarray}
\fl
    \left(
        \diag{ \hat{\bm{Q}}_{2x} }
        + 
        A^\top 
        \diag{ \hat{\bm{Q}}_{2z} } 
        A
    \right)^{-1}
    = \diag{ \hat{\bm{Q}}_{2x}^{-1} }
    \nonumber \\
    \hspace{-20pt}
    - 
    \diag{ \hat{\bm{Q}}_{2x}^{-1} }
    A^\top
    \left(
        \diag{ \hat{\bm{Q}}_{2z}^{-1} }
        +
        A
        \diag{ \hat{\bm{Q}}_{2x}^{-1} }
        A^\top
    \right)^{-1}
    A
    \diag{ \hat{\bm{Q}}_{2x}^{-1} }.
\end{eqnarray}
Because in high-dimensional statistics, the number of the samples in the data is often one or several orders of magnitude smaller than the number of the parameters, the computational cost is drastically reduced using this identity.

Second, for a real-world dataset with a small number of samples, VAMP trajectories can show large oscillations, which lead to poor convergence.
In such cases, introducing a small amount of damping factor $\eta_{\rm d}\in (0, 1]$ can improve the convergence of the algorithm. 
We suggest replacing line \ref{line:rVAMP moment-matching 2to1 1} and \ref{line:rVAMP moment-matching 2to1 2} with the damped versions:
\begin{eqnarray}
    \bm{h}_{1x}^{(t+1)} = \eta_{\rm d}\left(
        \frac{\hat{\bm{x}}_{2}^{(t)}}{\bm{\chi}_{2x}^{(t)}} - \bm{h}_{2x}^{(t)}
    \right) + (1-\eta_{\rm d})\bm{h}_{1x}^{(t)}
    ,\\
    \hat{\bm{Q}}_{1x}^{(t+1)} = \eta_{\rm d}\left(
        \frac{\bm{1}_N}{\bm{\chi}_{2x}^{(t)}} - \hat{\bm{Q}}_{2x}^{(t)}
    \right) + (1-\eta_{\rm d})\hat{\bm{Q}}_{1x}^{(t)}
    ,\\
    \hat{\bm{v}}_{1x}^{(t+1)} = \eta_{\rm d}\left(
        \frac{\bm{v}_{2x}^{(t)}}{ \left( \bm{\chi}_{2x}^{(t)} \right)^2} - \hat{\bm{v}}_{2x}^{(t)}
    \right) + (1-\eta_{\rm d})\hat{\bm{v}}_{1x}^{(t)}
    ,\\
        \bm{h}_{1z}^{(t+1)} = \eta_{\rm d}\left(
        \frac{\hat{\bm{z}}_{2}^{(t)}}{\bm{\chi}_{2z}^{(t)}} - \bm{h}_{2z}^{(t)}
    \right) + (1-\eta_{\rm d})\bm{h}_{1z}^{(t)}
    ,\\
    \hat{\bm{Q}}_{1z}^{(t+1)} = \eta_{\rm d}\left(
        \frac{\bm{1}_M}{\bm{\chi}_{2z}^{(t)}} - \hat{\bm{Q}}_{2z}^{(t)}
    \right) + (1-\eta_{\rm d})\hat{\bm{Q}}_{1z}^{(t)}
    ,\\
    \hat{\bm{v}}_{1z}^{(t+1)} = \eta_{\rm d}\left(
        \frac{\bm{v}_{2z}^{(t)}}{ \left( \bm{\chi}_{2z}^{(t)} \right)^2} - \hat{\bm{v}}_{2z}^{(t)}
    \right) + (1-\eta_{\rm d})\hat{\bm{v}}_{1z}^{(t)}.
\end{eqnarray}

Third, GLMs may require including an intercept term $z_0$ so that $y_\mu \sim p_{y|z}(y_\mu | z_0 + \bm{a}_\mu^\top\bm{x}_0)$. To incorporate the intercept term, we add an extra column in the feature matrix so that $A_{0,\mu} = 1,\mu=1,2,\dots,M$, and for this component we do not require any regularization term.

The last point regards how to obtain the selection probability for various values of the regularization strength $\gamma_0$. In practice, we are often interested in finding the selection probability not only for a single fixed $\gamma_0$, but also for the various regularization parameters $\gamma_0$ (as in Figure \ref{fig:compare_path_stability}). A reasonable approach is to begin with the largest $\gamma_0$. Then, we decrease $\gamma_0$ by a small amount and run rVAMP until convergence. Decreasing $\gamma_0$ again and using previous parameters at the fixed point as the initial conditions (\emph{warm start}), we then run rVAMP until convergence. Using this method, we can efficiently compute the selection probabilities over a grid of $\gamma_0$.

\section{Macroscopic analysis}
\label{sec:macro analysis}
The salient feature of the VAMP algorithms is that we can macroscopically analyze their convergence dynamics in a large system limit under specific assumptions on the distributions of the set of feature vectors. The derived dynamics are termed state evolution (SE). In this section, we derive SE for self-averaging rVAMP (SA rVAMP), which would describe the converging dynamics of rVAMP approximately. We also show that its fixed point is consistent with the replica symmetric solution obtained by the replica method, which is believed to be exact in the large system limit under appropriate conditions. Although the procedure of the replica method has not been justified mathematically yet, many studies have rigorously validated its conjectures in the last few decades, especially in Bayes optimal settings \cite{barbier2018mutual, barbier2019optimal, reeves2016replica, rangan2019vector}, and more recently in model-mismatched cases \cite{gerbelot2020asymptotic}. 

\subsection{Setup for the macroscopic analysis}
\label{subsection:setup for macro analysis}
For the theoretical analysis, we assume the actual data generation process as follows. First, the true parameter vector $\bm{x}_0$ and the response variables are generated as 
\begin{eqnarray}
    x_{0,i} \sim q_{x_0}(x_{0,i}), \quad i=1,2,\dots, N, \\
    y_\mu \sim q_{y|z}(y_\mu|\bm{a}_\mu^\top \bm{x}_0), \quad \mu=1,2,\dots,M.
\end{eqnarray}
Generally, the model used for the fitting and the actual generation model may be different $p_{y|z}\neq q_{y|z}$ or $e^{-\gamma|x|}\neq q_{x_0}$. Additionally, we assume that the feature matrix $A$ is drawn from the rotation-invariant random matrix ensembles, i.e. for the singular value decomposition $A=USV^\top, U\in\mathbb{R}^{M\times M}, S\in\mathbb{R}^{M\times N}, V\in\mathbb{R}^{N\times N}$, we assume that $U$ and $V$ are drawn from uniform distributions over $M\times M$ and $N\times N$ orthogonal matrices. 

We are interested in the large system limit where both of the numbers of data points and parameters diverge as $M,N\to\infty$ keeping the ratio $\alpha \equiv M/N\in(0,\infty)$. Because $U$ and $V$ are drawn independently from uniform distributions over $M\times M$ and $N\times N$ orthogonal matrices, for vectors $\bm{\omega}\in\mathbb{R}^N$ and $\bm{\phi}\in\mathbb{R}^M$, we expect that the empirical distributions of $V^\top \bm{\omega}$ and $U^\top\bm{\phi}$ converge to Gaussians with mean zero and variance $\|\bm{\omega}\|_2^2/N$ and $\|\bm{\phi}\|_2^2/M$ in this limit, respectively.

\subsection{Self-averaging rVAMP}
Our first interest is the convergence dynamics of rVAMP.
Unfortunately, directly investigating the dynamics of rVAMP is difficult because the time evolution of the empirical distributions of $\bm{h}_{1x}, \bm{h}_{1z}, \bm{h}_{2x}, \bm{h}_{2z}$ may not be described by a small number of statistics, although the dynamical-functional theory \cite{ccakmak2019memory, cakmak2017dynamical, martin1973statistical, eissfeller1992new} might give some insights for the raw rVAMP. To detour this difficulty approximately, we consider SA rVAMP, which eliminates the site dependence of the natural parameters in the approximate densities:
\begin{eqnarray}
    \hat{Q}_{1x,i}^{(t)} &= \hat{Q}_{1x}^{(t)}, \quad \hat{Q}_{2x,i}^{(t)} = \hat{Q}_{2x}^{(t)},\\
    \hat{v}_{1x,i}^{(t)} &= \hat{v}_{1x}^{(t)}, \quad \hat{v}_{2x,i}^{(t)} = \hat{v}_{2x}^{(t)},\\
    \hat{Q}_{1z,\mu}^{(t)} &= \hat{Q}_{1z}^{(t)}, \quad \hat{Q}_{2z,\mu}^{(t)} = \hat{Q}_{2z}^{(t)},\\
    \hat{v}_{1z,\mu}^{(t)} &= \hat{v}_{1z}^{(t)}, \quad \hat{v}_{2z,\mu}^{(t)} = \hat{v}_{2z}^{(t)}.
\end{eqnarray}
Eliminating the site dependence replaces the component-wise moment-matching conditions in (\ref{eq:second moment1})-(\ref{eq:second moment2}) with the macroscopic moment-matching conditions:
\begin{eqnarray}
    \frac{1}{N}\sum_{i=1}^N \int x_{\a,i}x_{\b,i}p_1^{(\beta)}d^n\bm{x}d^n\bm{z} &= \frac{1}{N}\sum_{i=1}^N \int x_{\a,i}x_{\b,i}p_2^{(\beta)}d^n\bm{x}d^n\bm{z} 
    \nonumber \\
    &= \frac{1}{N}\sum_{i=1}^N \int x_{\a,i}x_{\b,i}\tilde{p}_1^{(\beta)}\tilde{p}_2^{(\beta)}d^n\bm{x}d^n\bm{z},
    \\
    \frac{1}{M}\sum_{\mu=1}^M \int z_{\a,\mu}z_{\b,\mu}p_1^{(\beta)}d^n\bm{x}d^n\bm{z} &= \frac{1}{M}\sum_{\mu=1}^M \int z_{\a,\mu}z_{\b,\mu}p_2^{(\beta)}d^n\bm{x}d^n\bm{z} 
    \nonumber \\
    &= \frac{1}{M}\sum_{\mu=1}^M \int z_{\a,\mu}z_{\b,\mu}\tilde{p}_1^{(\beta)}\tilde{p}_2^{(\beta)}d^n\bm{x}d^n\bm{z}.
\end{eqnarray}
These modifications yield SA rVAMP described in algorithm \ref{algo:self-averaging rVAMP}. We will use it in the following analysis. 

\subsection{State evolution}
To derive the SE of SA rVAMP heuristically, we make the following assumptions following the literature \cite{takahashi2020macroscopic}. 

\vspace{2.0pt}
\noindent
\emph{\bf Assumption:} 
At each iteration $t=1,2,\dots ,T_{\rm iter}$, positive constants $\hat{m}_{kx}^{(t)}$, $\hat{m}_{kz}^{(t)}$, $\hat{\chi}_{kx}^{(t)}$, $\hat{\chi}_{kz}^{(t)}\in\mathbb{R}$, $(k=1,2)$ exist such that for the singular value decomposition $A=USV^\top$, 
\begin{eqnarray}
    \bm{h}_{1x}^{(t)} - \hat{m}_{1x}^{(t)}\bm{x}_0 &\doteq \sqrt{\hat{\chi}_{1x}^{(t)}}\bm{\xi}_{1x}^{(t)},
    \label{eq:assumption1}
    \\
    \bm{h}_{1z}^{(t)} - \hat{m}_{1z}^{(t)}\bm{z}_0 &\doteq \sqrt{\hat{\chi}_{1z}^{(t)}}\bm{\xi}_{1z}^{(t)}, 
    \label{eq:assumption2}
    \\
    V^\top (\bm{h}_{2x}^{(t)} - \hat{m}_{2x}^{(t)}\bm{x}_0) &\doteq \sqrt{\hat{\chi}_{2x}^{(t)}}\bm{\xi}_{2x}^{(t)}, 
    \label{eq:assumption3}
    \\
    U^\top (\bm{h}_{2z}^{(t)} - \hat{m}_{2z}^{(t)}\bm{z}_0) &\doteq \sqrt{\hat{\chi}_{2z}^{(t)}}\bm{\xi}_{2z}^{(t)},
    \label{eq:assumption4}
\end{eqnarray}
hold, where $\doteq$ denotes the equality of empirical distributions, $\bm{z}_0$ is $A\bm{x}_0$, and $\bm{\xi}_{kx}^{(t)}, \bm{\xi}_{kz}^{(t)}, (k=1,2, t=1,2,\dots ,T_{\rm iter})$ are mutually independent standard Gaussian variables. 

The equations (\ref{eq:assumption3}) and (\ref{eq:assumption4}) are expected from the mixing by randomly sampled orthogonal matrices $V^\top$ and $U^\top$.
The equations (\ref{eq:assumption1}) and (\ref{eq:assumption2}) are expected from the Onsager correction terms $-\bm{h}_{2x}^{(t)}, -\bm{h}_{2z}^{(t)}$ that appears in the moment-matching conditions in line \ref{line:moment-matching-sa-3}-\ref{line:moment-matching-sa-4}.

    \begin{algorithm}[p]
    \caption{self averaging rVAMP}
    \begin{algorithmic}[1]  \label{algo:self-averaging rVAMP}
    \REQUIRE{
        Denoising functions $g_{1x}, g_{1z}$ from (\ref{eq:denoiser1x}) and (\ref{eq:denoiser1z}), the features $A\in\mathbb{R}^{M\times N}$, the response variable $\bm{y}\in\mathbb{R}^M$, the convergence criterion $\epsilon_{\rm tol}$, and the maximum number of iterations $T_{\rm iter}$. 
    }
    \STATE{Select initial $\bm{h}_{1x}^{(1)}\in\mathbb{R}^N, \bm{h}_{1z}^{(1)}\in\mathbb{R}^M$, $\hat{Q}_{1x}^{(1)}, \hat{v}_{1x}^{(1)}, \hat{Q}_{1z}^{(1)}$, and $\hat{v}_{1z}^{(1)}\in[0,\infty)$.}
    \FOR{$t=1,2,\dots, T_{\rm iter}$}
        \STATE{// Factorized part}
        \vspace{2.0pt}
        \STATE{$
                \hat{\bm{x}}_{1}^{(t)} = \mathbb{E}_{\bm{\gamma}}[\int 
                    \bm{g}_{1x}(\bm{h}_{1x}^{(t)}, \hat{Q}_{1x}^{(t)}\bm{1}_N, \hat{v}_{1x}^{(t)}\bm{1}_N; \bm{\gamma}, \bm{\eta}_{x})
                D\bm{\eta}_{x}]
        $
        \label{line:factorized x1}
        }
        \vspace{2.0pt}
        \STATE{$
                \chi_{1x}^{(t)} = \langle
                    \mathbb{E}_{\bm{\gamma}}[\int 
                        \bm{g}_{1x}^\prime(\bm{h}_{1x}^{(t)}, \hat{Q}_{1x}^{(t)}\bm{1}_N, \hat{v}_{1x}^{(t)}\bm{1}_N; \bm{\gamma}, \bm{\eta}_{x})
                    D\bm{\eta}_{x}]
                \rangle
        $
        \label{line:factorized x2}
        }
        \vspace{2.0pt}
        \STATE{$
                v_{1x}^{(t)} = \langle
                    \mathbb{E}_{\bm{\gamma}}[\int 
                        \bm{g}_{1x}^2(\bm{h}_{1x}^{(t)}, \hat{Q}_{1x}^{(t)}\bm{1}_N, \hat{v}_{1x}^{(t)}\bm{1}_N; \bm{\gamma}, \bm{\eta}_{x})
                    D\bm{\eta}_{x}]
                    - (\hat{\bm{x}}_{1}^{(t)})^2
                \rangle
        $
        \label{line:factorized x3}
        }
        \STATE{$
                \hat{\bm{z}}_{1}^{(t)} = \mathbb{E}_{\bm{c}}[\int 
                    \bm{g}_{1z}(\bm{h}_{1z}^{(t)}, \hat{Q}_{1z}^{(t)}\bm{1}_M, \hat{v}_{1z}^{(t)}\bm{1}_M; \bm{c}, \bm{\eta}_{z}, \bm{y})
                D\bm{\eta}_{z}]
        $
        \label{line:factorized z1}
        }
        \vspace{2.0pt}
        \STATE{$
                \chi_{1z}^{(t)} = \langle
                    \mathbb{E}_{\bm{c}}[\int 
                        \bm{g}_{1z}^\prime(\bm{h}_{1z}^{(t)}, \hat{Q}_{1z}^{(t)}\bm{1}_M, \hat{v}_{1z}^{(t)}\bm{1}_M; \bm{c}, \bm{\eta}_{z}, \bm{y})
                    D\bm{\eta}_{z}]
                \rangle
        $
        \label{line:factorized z2}
        }
        \vspace{2.0pt}
        \STATE{$
                v_{1z}^{(t)} = \langle
                    \mathbb{E}_{\bm{c}}[\int 
                        \bm{g}_{1z}^2(\bm{h}_{1z}^{(t)}, \hat{Q}_{1z}^{(t)}\bm{1}_M, \hat{v}_{1z}^{(t)}\bm{1}_M; \bm{c}, \bm{\eta}_{z}, \bm{y})
                    D\bm{\eta}_{z}]
                    - (\hat{\bm{z}}_{1}^{(t)})^2
                \rangle
        $
        \label{line:factorized z3}
        }
        \vspace{1.5pt}
        \STATE{// Moment-matching ($1\to2$)}
        \vspace{2.0pt}
        \STATE{$
            \bm{h}_{2x}^{(t)} = \hat{\bm{x}}_{1}^{(t)}/(\chi_{1x}^{(t)}\bm{1}_N) - \bm{h}_{1x}^{(t)}
            ,\quad
            \hat{Q}_{2x}^{(t)} = \left(\chi_{1x}^{(t)}\right)^{-1} - \hat{Q}_{1x}^{(t)}
            ,\quad
            \hat{v}_{2x}^{(t)} = v_{1x}^{(t)} / \left( \chi_{1x}^{(t)} \right)^2 - \hat{v}_{1x}^{(t)}
        $\label{line:moment-matching-sa-1}}
        \STATE{$
                \bm{h}_{2z}^{(t)} = \hat{\bm{z}}_{1}^{(t)}/(\chi_{1z}^{(t)}\bm{1}_M) - \bm{h}_{1z}^{(t)}
                ,\quad
                \hat{Q}_{2z}^{(t)} = \left(\chi_{1z}^{(t)}\right)^{-1} - \hat{Q}_{1z}^{(t)}
                ,\quad
                \hat{v}_{2z}^{(t)} = v_{1z}^{(t)} / \left( \chi_{1z}^{(t)} \right)^2 - \hat{v}_{1z}^{(t)}
            $\label{line:moment-matching-sa-2}}
        \vspace{2.0pt}
        \STATE{// Gaussian part}
        \STATE{$
            X = \left(
                    \hat{Q}_{2x}^{(t)}I_N + \hat{Q}_{2z}^{(t)}A^\top A
                \right)^{-1}
        $
        \label{line:inverse}
        }
        \STATE{$
            \hat{\bm{x}}_2^{(t)} = X(\bm{h}_{2x}+ A^\top \bm{h}_{2z})
            , \quad
            \hat{\bm{z}}_2^{(t)} = A\hat{\bm{x}}_2^{(t)}
        $}
        \STATE{$
            \chi_{2x}^{(t)} = N^{-1}{\rm Tr}[X]
            , \quad
            \chi_{2z}^{(t)} = M^{-1}{\rm Tr}[AXA^\top]
        $}
        \vspace{1.5pt}
        \STATE{$
            \bm{v}_{2x}^{(t)} = N^{-1}{\rm Tr}\left[
                    X\left(
                        \diag{\hat{\bm{v}}_{2x}^{(t)}} + A^\top \diag{\hat{\bm{v}}_{2z}^{(t)}}A
                    \right)X
                \right]
        $}
        \vspace{1.5pt}
        \STATE{$
            \bm{v}_{2z}^{(t)} = N^{-1}{\rm Tr}\left[
                AX\left(
                    \diag{\hat{\bm{v}}_{2x}^{(t)}} + A^\top \diag{\hat{\bm{v}}_{2z}^{(t)}}A
                \right)XA^\top
            \right]
        $}
        \STATE{// Moment-matching ($2\to1$)}
        \STATE{$
            \bm{h}_{1x}^{(t+1)} = \hat{\bm{x}}_{2}^{(t)}/(\chi_{2x}^{(t)}\bm{1}_N) - \bm{h}_{2x}^{(t)}
            ,\,
            \hat{Q}_{1x}^{(t+1)} = \left(\chi_{2x}^{(t)}\right)^{-1} - \hat{Q}_{2x}^{(t)}
            ,\,
            \hat{v}_{1x}^{(t+1)} = v_{2x}^{(t)} / \left( \chi_{2x}^{(t)} \right)^2 - \hat{v}_{2x}^{(t)}
        $\label{line:moment-matching-sa-3}}
        \STATE{$
                \bm{h}_{1z}^{(t+1)} = \hat{\bm{z}}_{2}^{(t)}/(\chi_{2z}^{(t)}\bm{1}_M) - \bm{h}_{2z}^{(t)}
                ,\,
                \hat{Q}_{1z}^{(t+1)} = \left(\chi_{2z}^{(t)}\right)^{-1} - \hat{Q}_{2z}^{(t)}
                ,\,
                \hat{v}_{1z}^{(t+1)} = v_{2z}^{(t)} / \left( \chi_{2z}^{(t)} \right)^2 - \hat{v}_{2z}^{(t)}
            $\label{line:moment-matching-sa-4}}
        \IF{$\max\{\|\hat{\bm{x}}_1^{(t)} - \hat{\bm{x}}_2^{(t)}\|_2^2/N, \|\hat{\bm{z}}_1^{(t)} - \hat{\bm{z}}_2^{(t)}\|_2^2/M\} < \epsilon_{\rm tol}$}
            \STATE{$t\leftarrow T_{\rm iter}$}
            \STATE{break}
        \ENDIF
    \ENDFOR
    \RETURN{$\bm{h}_{1x}^{(T_{\rm iter})}, \hat{Q}_{1x}^{(T_{\rm iter})},\hat{v}_{1x}^{(T_{\rm iter})}$}
    \end{algorithmic}
    \end{algorithm}

To characterize macroscopic behavior of rVAMP, we introduce the following macroscopic order parameters for $t=1,2,\dots,T_{\rm iter}$:
\begin{eqnarray}
    m_{1x}^{(t)} = \frac{1}{N}\bm{x}_0^\top \hat{\bm{x}}_1^{(t)},
    \quad 
    m_{1z}^{(t)} = \frac{1}{M}\bm{z}_0^\top \hat{\bm{z}}_1^{(t)}, 
    \\
    q_{1x}^{(t)} = \frac{1}{N}\left\|
        \hat{\bm{x}}_{1}^{(t)}
    \right\|_2^2, 
    \quad
    q_{1z}^{(t)} = \frac{1}{M}\left\|
        \hat{\bm{z}}_{1}^{(t)}
    \right\|_2^2, 
    \\
    m_{2x}^{(t)} = \frac{1}{N}\bm{x}_0^\top \hat{\bm{x}}_2^{(t)},
    \quad 
    m_{2z}^{(t)} = \frac{1}{M}\bm{z}_0^\top \hat{\bm{z}}_2^{(t)}, 
    \\
    q_{2x}^{(t)} = \frac{1}{N}\left\|
        \hat{\bm{x}}_{2}^{(t)}
    \right\|_2^2, 
    \quad
    q_{2z}^{(t)} = \frac{1}{M}\left\|
        \hat{\bm{z}}_{2}^{(t)}
    \right\|_2^2, 
    \\
    T_x = \frac{1}{N}\left\|
        \bm{x}_0
    \right\|_2^2, \quad
    T_z = \frac{1}{M}\left\|
        \bm{z}_0
    \right\|_2^2.
\end{eqnarray}

These order parameters and the susceptibilities have limiting expressions in the limit $N\to\infty$. First, $q_{1x}^{(t)}$ can be written as 
\begin{eqnarray}
\fl
    q_{1x}^{(t)} \simeq \frac{1}{N}\sum_{i=1}^N
        \left(
            \mathbb{E}_{\gamma_i}\left[\int 
                g_{1x}(h_{1x,i}^{(t)}, \hat{Q}_{1x}^{(t)}, \hat{v}_{1x}^{(t)}; \gamma_i, \eta_{x,i})
                D\eta_{x,i}
            \right]
        \right)^2
    \nonumber \\
\fl
    \stackrel{N\to\infty}{\to} \mathbb{E}_{x_0}\left[\int 
        \left(
            \mathbb{E}_{\gamma}
            \left[
                \int 
                    g_{1x}(\hat{m}_{1x}^{(t)}x_0 + \sqrt{\hat{\chi}_{1x}^{(t)}}\xi_x, \hat{Q}_{1x}^{(t)}, \hat{v}_{1x}^{(t)}; \gamma, \eta_{x})
                D\eta_x
            \right]
        \right)^2
        D\xi_{x}
    \right].
\end{eqnarray}
Here, the summation is replaced with the average in the limit $N\to\infty$. The average $\mathbb{E}_\gamma[\dots]$ is with respect to the density $p(\gamma) = \delta(\gamma - \gamma_0)/2 + \delta(\gamma - 2\gamma_0)/2$.
Similar results can be obtained for $m_{1x}^{(t)}, m_{1z}^{(t)}, \chi_{1x}^{(t)}, v_{1x}^{(t)}, q_{1z}^{(t)}, \chi_{1z}^{(t)}$ and $v_{1z}^{(t)}$.
Next, for the singular value decomposition $A=USV^\top$, we denote by $\{\sqrt{\lambda}_i\}$ the diagonal elements of $S$. Then, $q_{2x}^{(t)}$ can be written as follows:
\begin{eqnarray}
\fl
    q_{2x}^{(t)} = \frac{1}{N}\sum_{i=1}^N 
    \left(
        \left(
            \hat{m}_{2x}^{(t)} + S^\top S\hat{m}_{2z}^{(t)}
        \right)(V^\top \bm{x}_0)
        +
        \left(
            \sqrt{\hat{\chi}_{2x}^{(t)}} \bm{\xi}_{2x}^{(t)} + \sqrt{ \hat{\chi}_{2z}^{(t)} }S^\top \bm{\xi}_{2z}^{(t)}
        \right)
    \right)^\top
    \nonumber \\
    \times 
    \left(
        \hat{Q}_{2x}^{(t)}I_N + S^\top S\hat{Q}_{2z}^{(t)}
    \right)^{-2}
    \nonumber \\
    \times
    \left(
        \left(
            \hat{m}_{2x}^{(t)} + S^\top S\hat{m}_{2z}^{(t)}
        \right)(V^\top \bm{x}_0)
        +
        \left(
            \sqrt{\hat{\chi}_{2x}^{(t)}} \bm{\xi}_{2x}^{(t)} + \sqrt{ \hat{\chi}_{2z}^{(t)} }S^\top \bm{\xi}_{2z}^{(t)}
        \right)
    \right)
    \nonumber \\
\fl
    \simeq \frac{1}{N}\sum_{i=1}^N 
        \frac{
            (
                \hat{m}_{2x}^{(t)} + \lambda_i\hat{m}_{2z}^{(t)}
            )^2(V^\top \bm{x}_0)_i^2
        }
        {
            (
                \hat{Q}_{2x}^{(t)} + \lambda_i\hat{Q}_{2z}^{(t)}
            )^{2}
        }
    + \frac{1}{N}\sum_{i=1}^N
    \frac{
        \hat{\chi}_{2x}^{(t)}\xi_{2x,i}^2 + \lambda_i \hat{\chi}_{2z}\xi_{2z,i}^2
    }{
        (
            \hat{Q}_{2x}^{(t)} + \lambda_i\hat{Q}_{2z}^{(t)}
        )^{2}
    }
    \nonumber \\
\fl
    \stackrel{N\to\infty}{\to}
    T_x
    \mathbb{E}_{\lambda}\left[\frac{
        (\hat{m}_{2x}^{(t)} + \lambda\hat{m}_{2z}^{(t)})^2
    }{
        (
            \hat{Q}_{2x}^{(t)} + \lambda\hat{Q}_{2z}^{(t)}
        )^2
    }\right]
    +
    \mathbb{E}_{\lambda}\left[\frac{
        (\hat{\chi}_{2x}^{(t)} + \lambda\hat{\chi}_{2z}^{(t)})
    }{
        (
            \hat{Q}_{2x}^{(t)} + \lambda\hat{Q}_{2z}^{(t)}
        )^2
    }\right],
\end{eqnarray}
where we used the independence between $\bm{\xi}_{2x}^{(t)}, \bm{\xi}_{2z}^{(t)}, \bm{x}_0$ and $\{\lambda_i\}$, and we denoted by $\mathbb{E}_\lambda[...]$ an average with respect to the limiting eigenvalue spectrum $\rho(\lambda)$ of $A^\top A$. The calculations for $m_{2x}^{(t)}, m_{2z}^{(t)}, \chi_{2x}^{(t)}, v_{2x}^{(t)}, q_{2z}^{(t)}, \chi_{2z}^{(t)}$ and $v_{2z}^{(t)}$ are similar. Finally, using the singular value decomposition $A=USV^\top$, $T_x$ and $T_z$ are written as 
\begin{eqnarray}
    T_x = \frac{1}{N}\sum_{i=1}^Nx_{0,i}^2 \stackrel{N\to\infty}{\to} \int x_0^2 q_{x_0}(x_0)dx_0, \\
    T_z \stackrel{N\to\infty}{\to} \mathbb{E}_{z_0}[z_0^2] =  \frac{\mathbb{E}_{\lambda}[\lambda]}{\alpha}T_x,
\end{eqnarray}
where the average of $z_0$ is taken with respect to a Gaussian measure
\begin{equation}
    \exp\left(
        -\frac{\hat{T}_z}{2}z_0^2
    \right)\sqrt{\frac{\hat{T}_z}{2\pi}}dz,\quad
    \hat{T}_z = \frac{\alpha}{\mathbb{E}_\lambda[\lambda]T_x},
    \label{eq:prior_z}
\end{equation}
based on the observation in \cite{kabashima2008inference}; for a vector $\bm{\omega}\in\mathbb{R}^N$ that is independent of $A$, the empirical distribution of $A\bm{\omega}$ is a Gaussian with mean zero and variance $\mathbb{E}_{\lambda}[\lambda]\|\bm{\omega}\|_2^2/(\alpha N)$ in the large system limit.

The moment-matching conditions also have the following limiting expressions. First, $\hat{m}_{2x}^{(t)}$ can be written as 
\begin{eqnarray}
    \hat{m}_{2x}^{(t)} 
    \stackrel{\rm (a)}{\to} 
    \frac{1}{\|x_0\|_2^2}\bm{x}_0^\top \bm{h}_{2x}^{(t)} 
    \nonumber \\
    \stackrel{\rm (b)}{=} 
    \frac{1}{\|x_0\|_2^2}\bm{x}_0^\top \left(
        \frac{\hat{\bm{x}}_1^{(t)}}{\chi_{1x}^{(t)}}
        -\bm{h}_{1x}^{(t)}
    \right) 
    \nonumber \\
    \stackrel{\rm (c)}{=} 
    \frac{m_{1x}^{(t)}}{T_x\chi_{1x}^{(t)}} - \hat{m}_{1x}^{(t)},
    \label{eq:update_m2x_hat}
\end{eqnarray}
where the limit (a) follows from the definition of $\hat{m}_{2x}^{(t)}$; (b) follows from the moment-matching condition of SA rVAMP; (c) follows from the definitions of $m_{1x}^{(t)}$ and $\hat{m}_{1x}^{(t)}$. For $\hat{\chi}_{2x}^{(t)}$, its update rule can be written as 
\begin{eqnarray}
    \hat{\chi}_{2x}^{(t)}
    \stackrel{\rm (a)}{\to}
    \frac{1}{N}\|\bm{h}_{2x}^{(t)} - \hat{m}_{2x}^{(t)}\bm{x}_0\|_2^2
    \nonumber 
    \\
    \stackrel{\rm (b)}{=}
    \frac{1}{N}
    \left\|
        \frac{\hat{\bm{x}}_1^{(t)}}{\chi_{1x}^{(t)}} 
        - 
        \frac{m_{1x}^{(t)}}{T_x\chi_{1x}^{(t)}}\bm{x}_0
        -
        \sqrt{\hat{\chi}_{1x}^{(t)}}\bm{\xi}_{1x}^{(t)}
    \right\|_2^2
    \nonumber \\
    \stackrel{\rm (c)}{=} 
    \frac{q_{1x}^{(t)}}{(\chi_{1x}^{(t)})^2}
    -\frac{(m_{1x}^{(t)})^2}{T_x(\chi_{1x}^{(t)})^2}
    + \chi_{1x}^{(t)}
    -2\frac{\sqrt{\hat{\chi}_{1x}^{(t)}}}{\chi_{1x}^{(t)}}\frac{1}{N}(\hat{\bm{x}}_{1x}^{(t)})^\top \bm{\xi}_{1x}^{(t)},
    \nonumber \\
    \stackrel{\rm (d)}{=} 
    \frac{q_{1x}^{(t)}}{(\chi_{1x}^{(t)})^2}
    -\frac{(m_{1x}^{(t)})^2}{T_x(\chi_{1x}^{(t)})^2}
    - \chi_{1x}^{(t)},
\end{eqnarray}
where (a) follows from the definition of $\hat{\chi}_{2x}^{(t)}$; (b) follows from the moment-matching condition of SA rVAMP and the assumption 2; (c) uses the independence between $\bm{x}_0$ and $\bm{\xi}_{1x}^{(t)}$, and the definition of $m_{1x}^{(t)}$; (d) can be obtained from the following integration by parts according to 
\begin{eqnarray}
\fl
    \frac{1}{N}(\hat{\bm{x}}_{1x}^{(t)})^\top \bm{\xi}_{1x}^{(t)}
    &\to
    \mathbb{E}_{x_0}\left[\int 
            \mathbb{E}_{\gamma}
            \left[
                \int 
                    g_{1x}(\hat{m}_{1x}^{(t)}x_0 + \sqrt{\hat{\chi}_{1x}^{(t)}}\xi_x, \hat{Q}_{1x}^{(t)}, \hat{v}_{1x}^{(t)}; \lambda, \eta_{x})
                D\eta_x
            \right]
        \xi_x
        D\xi_{x}
    \right]
    \nonumber \\
\fl 
    &=  \sqrt{\hat{\chi}_{1x}^{(t)}}\mathbb{E}_{x_0}\left[\int 
            \mathbb{E}_{\gamma}
            \left[
                \int 
                    g_{1x}'(\hat{m}_{1x}^{(t)}x_0 + \sqrt{\hat{\chi}_{1x}^{(t)}}\xi_x, \hat{Q}_{1x}^{(t)}, \hat{v}_{1x}^{(t)}; \lambda, \eta_{x})
                D\eta_x
            \right]
        D\xi_{x}
    \right]
    \nonumber \\
    &=\sqrt{\hat{\chi}_{1x}^{(t)}}\chi_{1x}^{(t)}.
\end{eqnarray}
Similarly, $\hat{m}_{1x}^{(t+1)}$ and $\hat{\chi}_{1x}^{(t+1)}$ are obtained as follows. For $\hat{m}_{1x}^{(t+1)}$, its update rule is derived exactly same way as in (\ref{eq:update_m2x_hat}). For $\hat{\chi}_{1x}^{(t)}$, 
\begin{eqnarray}
    \hat{\chi}_{1x}^{(t)} 
    \stackrel{\rm (a)}{\to}
    \frac{1}{N}\|\bm{h}_{1x}^{(t+1)} - \hat{m}_{1x}^{(t+1)}\bm{x}_0\|_2^2
    \nonumber \\
    \stackrel{\rm (b)}{=}
    \frac{1}{N}\left\|
        \frac{\hat{\bm{x}}_2}{\chi_{2x}^{(t)}} - \frac{m_{2x}^{(t)}}{T_x\chi_{2x}^{(t)}}\bm{x}_0 - \sqrt{\hat{\chi}_{2x}^{(t)}}V\bm{\xi}_{2x}^{(t)}
    \right\|_2^2
    \nonumber \\
    \stackrel{\rm (c)}{=}
    \frac{q_{2x}^{(t)}}{(\chi_{2x}^{(t)})^2} - \frac{(m_{2x}^{(t)})^2}{T_x(\chi_{2x}^{(t)})^2} 
    +
    \hat{\chi}_{2x}^{(t)}
    -2\frac{\sqrt{\hat{\chi}_{2x}^{(t)}}}{\chi_{2x}^{(t)}} \frac{1}{N}(V^\top  \hat{\bm{x}}_2^{(t)})^\top \bm{\xi}_{2x}^{(t)}
    \nonumber \\
    \stackrel{\rm (d)}{=}
    \frac{q_{2x}^{(t)}}{(\chi_{2x}^{(t)})^2} - \frac{(m_{2x}^{(t)})^2}{T_x(\chi_{2x}^{(t)})^2} -\hat{\chi}_{2x}^{(t)},
\end{eqnarray}
where (a) follows from the definition of $\hat{\chi}_{1x}^{(t+1)}$; (b) follows from the moment-matching condition and the assumption 2; (c) uses the independence between $V^\top\bm{x}_0$ and $\bm{\xi}_{2x}^{(t)}$, and the definition of $m_{2x}^{(t)}$; (d) can be obtained from the independence between $V^\top\bm{x}_0, S^\top \bm{\xi}_{2z}^{(t)}$ and  $\bm{\xi}_{2x}^{(t)}$:
\begin{eqnarray}
\fl
    \frac{1}{N}(V^\top \hat{\bm{x}}_2^{(t)})^\top \bm{\xi}_{2x}^{(t)}
    = \frac{1}{N}\sum_{i=1}^N \frac{
        \left(
            (\hat{m}_{2x}^{(t)} + \lambda_i\hat{m}_{2z}^{(t)})[V^\top\bm{x}_0]_i
            + \sqrt{\hat{\chi}_{2x}^{(t)}} \xi_{2x,i}
            + \sqrt{\hat{\chi}_{2z}^{(t)}}[S^\top (\bm{\xi}_{2z}^{(t)})]_i
        \right)\xi_{2x,i}^{(t)}
    }{
        \hat{Q}_{2x}^{(t)} + \lambda_i\hat{Q}_{2z}^{(t)}
    }
    \nonumber \\
    \to \sqrt{\hat{\chi}_{2x}^{(t)}}\mathbb{E}_\lambda\left[\frac{1}{\hat{Q}_{2x}^{(t)} + \lambda \hat{Q}_{2z}^{(t)}}\right]\int \xi_{2x}^2 D\xi_{2x}
    \nonumber\\
    = \sqrt{\hat{\chi}_{2x}^{(t)}} \chi_{2x}^{(t)}.
\end{eqnarray}
Similar results can be obtained for $\hat{m}_{2z}^{(t)}, \hat{\chi}_{2z}^{(t)}, \hat{m}_{1z}^{(t+1)}$ and $\hat{\chi}_{1z}^{(t+1)}$.

The above observations yield the SE of SA rVAMP as follows:

\noindent
\emph{Initialization:} Select initial $\hat{m}_{1x}^{(1)}, \hat{\chi}_{1x}^{(1)}, \hat{Q}_{1x}^{(1)}, \hat{v}_{1x}^{(1)}, \hat{m}_{1z}^{(1)}, \hat{\chi}_{1z}^{(1)}, \hat{Q}_{1z}^{(1)}$, and $\hat{v}_{1z}^{(1)}\in[0,\infty)$.

\noindent
\emph{Iteration:} For $t=1,2,\dots,T_{\rm iter}$, update the parameters as follows:

\emph{Factorized part:}
\begin{eqnarray}
\fl
    q_{1x}^{(t)} = \mathbb{E}_{x_0}\left[\int 
        \left(
            \mathbb{E}_{\gamma}
            \left[
                \int 
                    g_{1x}(\hat{m}_{1x}^{(t)}x_0 + \sqrt{\hat{\chi}_{1x}^{(t)}}\xi_x, \hat{Q}_{1x}^{(t)}, \hat{v}_{1x}^{(t)}; \gamma, \eta_{x})
                D\eta_x
            \right]
        \right)^2
        D\xi_{x}
    \right],
    \label{eq:se_factorized_1}
    \\
\fl
    \chi_{1x}^{(t)} = \mathbb{E}_{x_0}\left[\int 
            \mathbb{E}_{\gamma}
            \left[
                \int 
                    g_{1x}^\prime(\hat{m}_{1x}^{(t)}x_0 + \sqrt{\hat{\chi}_{1x}^{(t)}}\xi_x, \hat{Q}_{1x}^{(t)}, \hat{v}_{1x}^{(t)}; \gamma, \eta_{x})
                D\eta_x
            \right]
        D\xi_{x}
    \right],
    \label{eq:se_factorized_2}
    \\
\fl 
    v_{1x}^{(t)} = \mathbb{E}_{x_0}\left[\int 
            \mathbb{E}_{\gamma}
            \left[
                \int 
                    g_{1x}^2(\hat{m}_{1x}^{(t)}x_0 + \sqrt{\hat{\chi}_{1x}^{(t)}}\xi_x, \hat{Q}_{1x}^{(t)}, \hat{v}_{1x}^{(t)}; \gamma, \eta_{x})
                D\eta_x
            \right]
        D\xi_{x}
    \right]
    \nonumber \\
    \hspace{-20pt}
    -
    \mathbb{E}_{x_0}\left[\int 
        \left(
            \mathbb{E}_{\gamma}
            \left[
                \int 
                    g_{1x}(\hat{m}_{1x}^{(t)}x_0 + \sqrt{\hat{\chi}_{1x}^{(t)}}\xi_x, \hat{Q}_{1x}^{(t)}, \hat{v}_{1x}^{(t)}; \gamma, \eta_{x})
                D\eta_x
            \right]
        \right)^2
        D\xi_{x}
    \right],
    \label{eq:se_factorized_3}
    \\
\fl
    m_{1x}^{(t)}=  \mathbb{E}_{x_0}\left[\int 
        x_0
            \mathbb{E}_{\gamma}
            \left[
                \int 
                    g_{1x}(\hat{m}_{1x}^{(t)}x_0 + \sqrt{\hat{\chi}_{1x}^{(t)}}\xi_x, \hat{Q}_{1x}^{(t)}, \hat{v}_{1x}^{(t)}; \gamma, \eta_{x})
                D\eta_x
            \right]
        D\xi_{x}
    \right],
    \label{eq:se_factorized_4}
    \\
\fl
    q_{1z}^{(t)} = \mathbb{E}_{z_0}\left[\int 
        \left(
            \mathbb{E}_{c}
            \left[
                \int 
                    g_{1z}(\hat{m}_{1z}^{(t)}z_0 + \sqrt{\hat{\chi}_{1z}^{(t)}}\xi_z, \hat{Q}_{1z}^{(t)}, \hat{v}_{1z}^{(t)}; c, \eta_{z}, y)
                D\eta_x
            \right]
        \right)^2
    \right.
    \nonumber \\
    \left.
        \times q_{y|z}(y|z_0)dy
        D\xi_{z}
    \right]
    ,
    \label{eq:se_factorized_5}
    \\
\fl
    \chi_{1z}^{(t)} = \mathbb{E}_{z_0}\left[\int 
            \mathbb{E}_{c}
            \left[
                \int 
                    g_{1z}^\prime(\hat{m}_{1z}^{(t)}z_0 + \sqrt{\hat{\chi}_{1z}^{(t)}}\xi_z, \hat{Q}_{1z}^{(t)}, \hat{v}_{1z}^{(t)}; c, \eta_{z}, y)
                D\eta_z
            \right]
        \right.
        \nonumber \\
        \left.
            \times q_{y|z}(y|z_0)dy
        D\xi_{z}
    \right],
    \label{eq:se_factorized_6}
    \\
\fl
    v_{1z}^{(t)} = \mathbb{E}_{z_0}\left[\int 
            \mathbb{E}_{c}
            \left[
                \int 
                    g_{1z}^2(\hat{m}_{1z}^{(t)}z_0 + \sqrt{\hat{\chi}_{1z}^{(t)}}\xi_z, \hat{Q}_{1z}^{(t)}, \hat{v}_{1z}^{(t)}; c, \eta_{z}, y)
                D\eta_z
            \right]
        q_{y|z}(y|z_0)dy
        D\xi_{z}
    \right]
    \nonumber \\
    \hspace{-50pt}
    - 
    \mathbb{E}_{z_0}\left[\int 
        \left(
            \mathbb{E}_{c}
            \left[
                \int 
                    g_{1z}(\hat{m}_{1z}z_0 + \sqrt{\hat{\chi}_{1z}^{(t)}}\xi_z, \hat{Q}_{1z}^{(t)}, \hat{v}_{1z}^{(t)}; c, \eta_{z}, y)
                D\eta_z
            \right]
        \right)^2
    \right.
    \nonumber \\
    \left.
        \times q_{y|z}(y|z_0)dy
        D\xi_{z}
    \right],
    \label{eq:se_factorized_7}
    \\
\fl
    m_{1z}^{(t)} = \mathbb{E}_{z_0}\left[\int 
            z_0
            \mathbb{E}_{c}
            \left[
                \int 
                    g_{1z}(\hat{m}_{1z}z_0 + \sqrt{\hat{\chi}_{1z}^{(t)}}\xi_z, \hat{Q}_{1z}^{(t)}, \hat{v}_{1z}^{(t)}; c, \eta_{z}, y)
                D\eta_z
            \right]
    \right.
    \nonumber \\
    \left.
        \times q_{y|z}(y|z_0)dy
        D\xi_{z}
    \right].
    \label{eq:se_factorized_8}
\end{eqnarray}

\emph{Moment-matching:}
\begin{eqnarray}
    \hat{Q}_{2x}^{(t)} = \frac{1}{\chi_{1x}^{(t)}} - \hat{Q}_{1x}^{(t)}, 
    \quad
    \hat{Q}_{2z}^{(t)} = \frac{1}{\chi_{1z}^{(t)}} - \hat{Q}_{1z}^{(t)},
    \label{eq:se_moment_matching_f_to_g_1}
    \\
    \hat{v}_{2x}^{(t)} = \frac{v_{1x}^{(t)}}{(\chi_{1x}^{(t)})^2} - \hat{v}_{1x}^{(t)},
    \quad
    \hat{v}_{2z}^{(t)} = \frac{v_{1z}^{(t)}}{(\chi_{1z}^{(t)})^2} - \hat{v}_{1z}^{(t)},
    \label{eq:se_moment_matching_f_to_g_2}
    \\
    \hat{m}_{2x}^{(t)} = \frac{m_{1x}^{(t)}}{T_x\chi_{1x}^{(t)}} - \hat{m}_{1x}^{(t)},
    \quad
    \hat{m}_{2z}^{(t)} = \frac{m_{1z}^{(t)}}{T_z\chi_{1z}^{(t)}} - \hat{m}_{1z}^{(t)},
    \label{eq:se_moment_matching_f_to_g_3}
    \\
    \hat{\chi}_{2x}^{(t)} = \frac{q_{1x}^{(t)}}{(\chi_{1x}^{(t)})^2} - \frac{(m_{1x}^{(t)})^2}{T_x(\chi_{1x}^{(t)})^2} - \hat{\chi}_{1x}^{(t)},
    \quad
    \hat{\chi}_{2z}^{(t)} = \frac{q_{1z}^{(t)}}{(\chi_{1z}^{(t)})^2} - \frac{(m_{1z}^{(t)})^2}{T_z(\chi_{1z}^{(t)})^2} - \hat{\chi}_{1z}^{(t)}.
    \label{eq:se_moment_matching_f_to_g_4}
\end{eqnarray}

\emph{Gaussian part:}
\begin{eqnarray}
    q_{2x}^{(t)} = T_x
    \mathbb{E}_{\lambda}\left[\frac{
        (\hat{m}_{2x}^{(t)} + \lambda\hat{m}_{2z}^{(t)})^2
    }{
        (
            \hat{Q}_{2x}^{(t)} + \lambda\hat{Q}_{2z}^{(t)}
        )^2
    }\right]
    +
    \mathbb{E}_{\lambda}\left[\frac{
        (\hat{\chi}_{2x}^{(t)} + \lambda\hat{\chi}_{2z}^{(t)})
    }{
        (
            \hat{Q}_{2x}^{(t)} + \lambda\hat{Q}_{2z}^{(t)}
        )^2
    }\right],
    \label{eq:se_gaussian_1}
    \\
    \chi_{2x}^{(t)} = \mathbb{E}_{\lambda}\left[
        \frac{1}{\hat{Q}_{2x}^{(t)} + \lambda\hat{Q}_{2z}^{(t)}}
    \right],
    \label{eq:se_gaussian_2}
    \\
    v_{2x}^{(t)} = \mathbb{E}_{\lambda}\left[
        \frac{
            \hat{v}_{2x}^{(t)} + \lambda\hat{v}_{2z}^{(t)}
        }{
            (
                \hat{Q}_{2x}^{(t)} + \lambda\hat{Q}_{2z}^{(t)}
            )^2
        }
    \right],
    \label{eq:se_gaussian_3}
    \\
    m_{2x}^{(t)}=
    T_x\mathbb{E}_{\lambda}\left[
        \frac{
            \hat{m}_{2x}^{(t)} + \lambda \hat{m}_{2z}^{(t)}
        }{
            \hat{Q}_{2x}^{(t)} + \lambda\hat{Q}_{2z}^{(t)}
        } 
    \right],
    \label{eq:se_gaussian_4}
    \\
    q_{2z}^{(t)}= 
    \frac{T_x}{\alpha}
    \mathbb{E}_{\lambda}\left[\frac{
        \lambda(\hat{m}_{2x}^{(t)} + \lambda\hat{m}_{2z}^{(t)})^2
    }{
        (
            \hat{Q}_{2x}^{(t)} + \lambda\hat{Q}_{2z}^{(t)}
        )^2
    }\right]
    +
    \mathbb{E}_{\lambda}\left[\frac{
        \lambda(\hat{\chi}_{2x}^{(t)} + \lambda\hat{\chi}_{2z}^{(t)})
    }{
        (
            \hat{Q}_{2x}^{(t)} + \lambda\hat{Q}_{2z}^{(t)}
        )^2
    }\right],
    \label{eq:se_gaussian_5}
    \\
    \chi_{2z} = 
    \frac{1}{\alpha}\mathbb{E}_{\lambda}\left[
        \frac{
            \lambda
        }{
            \hat{Q}_{2x}^{(t)} + \lambda\hat{Q}_{2z}^{(t)}
        }
    \right],
    \label{eq:se_gaussian_6}
    \\
    v_{2z}^{(t)} = \frac{1}{\alpha}\mathbb{E}_{\lambda}\left[
        \frac{
            \lambda(\hat{v}_{2x}^{(t)} + \lambda\hat{v}_{2z}^{(t)})
        }{
            (
                \hat{Q}_{2x}^{(t)} + \lambda\hat{Q}_{2z}^{(t)}
            )^2
        }
    \right],
    \label{eq:se_gaussian_7}
    \\
    m_{2z}^{(t)}=
    \frac{T_x}{\alpha}
    \mathbb{E}_{\lambda}\left[
        \frac{
            \lambda(\hat{m}_{2x}^{(t)} + \lambda \hat{m}_{2z}^{(t)})
        }{
            \hat{Q}_{2x}^{(t)} + \lambda\hat{Q}_{2z}^{(t)}
        } 
    \right].
    \label{eq:se_gaussian_8}
\end{eqnarray}

\emph{Moment-matching:}
\begin{eqnarray}
    \hat{Q}_{1x}^{(t+1)} = \frac{1}{\chi_{2x}^{(t)}} - \hat{Q}_{2x}^{(t)}, 
    \quad
    \hat{Q}_{1z}^{(t+1)} = \frac{1}{\chi_{2z}^{(t)}} - \hat{Q}_{2z}^{(t)},
    \label{eq:se_moment_matching_g_to_f_1}
    \\
    \hat{v}_{1x}^{(t+1)} = \frac{v_{2x}^{(t)}}{(\chi_{2x}^{(t)})^2} - \hat{v}_{1x}^{(t)},
    \quad
    \hat{v}_{1z}^{(t+1)} = \frac{v_{2z}^{(t)}}{(\chi_{2z}^{(t)})^2} - \hat{v}_{1z}^{(t)},
    \label{eq:se_moment_matching_g_to_f_2}
    \\
    \hat{m}_{1x}^{(t+1)} = \frac{m_{2x}^{(t)}}{T_x\chi_{2x}^{(t)}} - \hat{m}_{1x}^{(t)},
    \quad
    \hat{m}_{1z}^{(t+1)} = \frac{m_{2z}^{(t)}}{T_z\chi_{2z}^{(t)}} - \hat{m}_{1z}^{(t)},
    \label{eq:se_moment_matching_g_to_f_3}
    \\
    \hspace{-10pt}
    \hat{\chi}_{1x}^{(t+1)} = \frac{q_{2x}^{(t)}}{(\chi_{2x}^{(t)})^2} - \frac{(m_{2x}^{(t)})^2}{T_x(\chi_{2x}^{(t)})^2} - \hat{\chi}_{2x}^{(t)},
    \,
    \hat{\chi}_{1z}^{(t+1)} = \frac{q_{2z}^{(t)}}{(\chi_{2z}^{(t)})^2} - \frac{(m_{2z}^{(t)})^2}{T_z(\chi_{2z}^{(t)})^2} - \hat{\chi}_{2z}^{(t)},
    \label{eq:se_moment_matching_g_to_f_4}
\end{eqnarray}
where $\mathbb{E}_c[\dots]$ is the average with respect to the probability function $p(c)=e^{-1}/c!,c=0,1,\dots$.

At the fixed point, $q_{1x}^{(t)}=q_{2x}^{(t)}, \chi_{1x}^{(t)}=\chi_{2x}^{(t)}, v_{1x}^{(t)}=v_{2x}^{(t)}$, and $m_{1x}^{(t)}=m_{2x}^{(t)}$ are approximate values of the following quantities:
\begin{eqnarray}
    q_x \simeq \lim_{\beta\to\infty,N\to\infty}\frac{1}{N}\left\|
        \mathbb{E}_{\bm{c},\bm{\gamma}}\left[\int 
            \bm{x} p^{(\beta)}(\bm{x}, \bm{z};\bm{c}, \bm{\gamma}, D)
            d\bm{x}d\bm{z}
        \right]
    \right\|_2^2,
    \label{eq:approx_qx}
    \\
    \chi_x \simeq \lim_{\beta\to\infty,N\to\infty}\frac{\beta}{N}\mathbb{E}_{\bm{c},\bm{\gamma}}\left[
            \int 
                \|\bm{x}\|_2^2
            p^{(\beta)}(\bm{x},\bm{z};\bm{c},\bm{\gamma},D)d\bm{x}d\bm{z}
        \right.
        \nonumber \\
       \hspace{120pt} \left.
            -
            \left\|
                \int
                    \bm{x}
                p^{(\beta)}(\bm{x},\bm{z};\bm{c},\bm{\gamma},D)d\bm{x}d\bm{z}
            \right\|_2^2
    \right],
    \label{eq:approx_chix}
    \\
    v_x \simeq \lim_{\beta\to\infty,N\to\infty}\frac{1}{N}
    \left(
        \mathbb{E}_{\bm{c},\bm{\gamma}}\left[
            \left\|
                    \int
                        \bm{x}
                    p^{(\beta)}(\bm{x},\bm{z};\bm{c},\bm{\gamma},D)d\bm{x}d\bm{z}
                \right\|_2^2
        \right]
    \right.
        \nonumber \\
        \hspace{120pt}-
    \left.
        \left\|
            \mathbb{E}_{\bm{c},\bm{\gamma}}\left[
                    \int
                        \bm{x}
                    p^{(\beta)}(\bm{x},\bm{z};\bm{c},\bm{\gamma},D)d\bm{x}d\bm{z}
            \right]
        \right\|_2^2
    \right)
    \label{eq:approx_vx}
    \\
    m_x \simeq \lim_{\beta\to\infty,N\to\infty}\mathbb{E}_{\bm{c},\bm{\gamma}}\left[
        \bm{x}_0^\top 
        \int
            \bm{x}
        p^{(\beta)}(\bm{x},\bm{z};\bm{c},\bm{\gamma},D)d\bm{x}d\bm{z}
    \right].
    \label{eq:approx_mx}
\end{eqnarray}
A similar interpretation is also possible for $q_{1z}^{(t)}=q_{2z}^{(t)}, \chi_{1z}^{(t)}=\chi_{2z}^{(t)}, v_{1z}^{(t)}=v_{2z}^{(t)}$, and $m_{1z}^{(t)}=m_{2z}^{(t)}$.

\subsection{Replica analysis}
Generally, typical values of the macroscopic order parameters introduced in the last section can be obtained by calculating the Helmholtz free energy $f$ using the replica method \cite{mezard1987spin}:
\begin{eqnarray}
    f=\mathbb{E}_{D}\left[f(D)\right] &\equiv -\lim_{
        N,
        \beta\to\infty,
        n\to0
    }\frac{1}{Nn\beta}\mathbb{E}_{D}[\log \Xi_n(D)]
    \\
    &= -\lim_{N,\beta\to\infty \atop n,\tilde{l}\to0}
    \frac{1}{Nn\tilde{l}\beta}\mathbb{E}_{D}\left[\Xi_n(D)^{\tilde{l}}\right].
\end{eqnarray}
Although the above formula contains the nested replicas, its replica symmetric computation is formally analogous to the standard 1-step replica symmetry breaking (1-RSB) computation by treating $\tilde{l}$ as the Parisi's breaking parameter. Because the 1-RSB computation was already described in appendix C of reference \cite{takahashi2020macroscopic}, we only show the final result. By rescaling the replica number as $\tilde{l}=l/\beta$, we obtain the following expression:
\begin{eqnarray}
\fl
    f &= - \lim_{\beta\to\infty, l\to 0}\mathop{\rm extr}_{m_x, q_x, v_x, \chi_x, \atop
        m_z, q_z, v_z, \chi_z}\left[g_{\rm F} + g_{\rm G} - g_{\rm S}\right],
            \\
\fl
    g_{\rm F} &= \mathop{\rm extr}_{
              \hat{m}_{1x}, \hat{\chi}_{1x}, \hat{v}_{1x}, \hat{Q}_{1x}, \atop
              \hat{m}_{1z}, \hat{\chi}_{1z}, \hat{v}_{1z}, \hat{Q}_{1z}
        }
        \left[
            -m_x\hat{m}_{1x}
            +\frac{1}{2}\left(
                q_x + v_x + \frac{\chi_x}{\beta}
            \right)\hat{Q}_{1x}
            - \frac{l}{2}\left(
                    (q_x+v_x)(\hat{\chi}_{1x} + \hat{v}_{1x}
                ) - q_x \hat{\chi}_{1x}
            \right)
        \right.    
        \nonumber \\ 
\fl
        &\left.
            -\frac{1}{2}\chi_x (\hat{\chi}_{1x} + \hat{v}_{1x})
            -\alpha m_z \hat{m}_{1z} 
            + \frac{\alpha}{2}\left(
                q_z + v_z + \frac{\chi_z}{\beta}
            \right)\hat{Q}_{1z}
            -\frac{l\alpha}{2}((q_z + v_z)(\hat{\chi}_{1z}+\hat{v}_{1z})- q_z\hat{\chi}_{1z}) 
        \right.
        \nonumber \\
\fl
        &\left.
            -\frac{\alpha}{2}\chi_z(\hat{\chi}_{1z} + \hat{v}_{1z})
            + \frac{1}{l}\int\left\{
                \log 
                \mathbb{E}_{\gamma}\left[\int
                    e^{l\phi_x^{(\beta)}}
                D\eta_x\right]
            \right\}q_{x_0}(x_0)dx_0D\xi_x
        \right.
        \nonumber \\
\fl
        &\left.
            + 
            \frac{1}{l}\int\left\{
                \log \mathbb{E}_c\left[
                    \int 
                        e^{l\phi_z^{(\beta)}}
                    D\eta_z
                \right]
            \right\}\sqrt{\frac{\hat{T}_z}{2\pi }}e^{-\frac{\hat{T}_z}{2}z_0^2}q_{y|z}(y|z_0) dz_0D\xi_z dy
        \right],
        \label{eq:g_F_1rsb} 
    \\
\fl
    g_{G} &= 
    \mathop{\rm extr}_{
              \hat{m}_{2x}, \hat{\chi}_{2x}, \hat{v}_{2x}, \hat{Q}_{2x}, \atop
              \hat{m}_{2z}, \hat{\chi}_{2z}, \hat{v}_{2z}, \hat{Q}_{2z}
        }
        \left[
            -m_x\hat{m}_{2x}
                +\frac{1}{2}\left(
                    q_x + v_x + \frac{\chi_x}{\beta}
                \right)\hat{Q}_{2x}
            - \frac{l}{2}((q_x+v_x)(\hat{\chi}_{2x} + \hat{v}_{2x}) - q_x \hat{\chi}_{2x})
        \right.
        \nonumber \\
\fl
        &\left.
            -\frac{1}{2}\chi_x (\hat{\chi}_{2x} + \hat{v}_{2x})
            -\alpha m_z \hat{m}_{2z} 
            + \frac{\alpha}{2}\left(
                q_z + v_z + \frac{\chi_z}{\beta}
            \right)\hat{Q}_{2z}
            -\frac{\alpha l}{2}((q_z + v_z)(\hat{\chi}_{2z}+\hat{v}_{2z})- q_z\hat{\chi}_{2z}) 
        \right.
        \nonumber \\
\fl
        &\left.     
            -\frac{\alpha}{2}\chi_z(\hat{\chi}_{2z} + \hat{v}_{2z})
            -\frac{1}{2}\left(\frac{1}{\beta} - \frac{1}{l}\right)\mathbb{E}_\lambda\left[
                \log \left(
                    \hat{Q}_{2x} + \lambda\hat{Q}_{2z}
                    \right)
            \right]
        \right.
        \nonumber \\
\fl        
        &\left.
            -\frac{1}{2l}\mathbb{E}_\lambda\left[
                \log \left(
                    \hat{Q}_{2x} + \lambda\hat{Q}_{2z} -l (\hat{v}_{2x} + \lambda\hat{v}_{2z})
                    \right)
            \right]
        \right.
        \nonumber \\
\fl
        &\left.
            +\frac{1}{2}\mathbb{E}_\lambda\left[
                \frac{
                    \hat{\chi}_{2x} + \lambda\hat{\chi}_{2z}
                }{
                    \hat{Q}_{2x} + \lambda\hat{Q}_{2z} -l (\hat{v}_{2x} + \lambda\hat{v}_{2z})
                }
            \right]
            +\frac{T_x}{2}\mathbb{E}_{\lambda}\left[
                \frac{
                    \left(\hat{m}_{2x} + \lambda\hat{m}_{2z}\right)^2
                }{
                    \hat{Q}_{2x} + \lambda\hat{Q}_{2z} -l (\hat{v}_{2x} + \lambda\hat{v}_{2z})
                }
            \right]
        \right] 
        \label{eq:g_G_1rsb},
        \\
\fl
        g_{\rm S} &= 
        \frac{1}{2}\left(\frac{1}{\beta} - \frac{1}{l}\right)\log \chi_x + \frac{1}{2l}\log (\chi_x + lv_x)
        + \frac{1}{2}\frac{q_x}{\chi_x + lv_x} - \frac{1}{2}\frac{m_x^2}{T_x(\chi_x + lv_x)}
        \nonumber \\
\fl
        & + \frac{\alpha}{2}\left(\frac{1}{\beta} - \frac{1}{l}\right)\log \chi_z + \frac{\alpha}{2l}\log (\chi_z + lv_z)
        + \frac{\alpha}{2}\frac{q_x}{\chi_z + lv_z} - \frac{\alpha}{2}\frac{m_z^2}{T_z(\chi_z + lv_z)},
        \label{eq:g_S_1rsb}
\end{eqnarray}
where 
\begin{eqnarray}
    \phi_x^{(\beta)} = \frac{1}{\beta}\log \int e^{
        -\beta\frac{\hat{Q}_{1x}}{2}x^2 
        + \beta (\hat{m}_{1x}x_0 + \sqrt{\hat{\chi}_{1x}}\xi_{x} + \sqrt{\hat{v}_{1x}}\eta_x)x
        - \beta \gamma|x|
    }dx,
    \\
    \phi_z^{(\beta)} = \frac{1}{\beta}\log \int e^{
        -\beta\frac{\hat{Q}_{1z}}{2}z^2 
        + \beta (\hat{m}_{1z}z_0 + \sqrt{\hat{\chi}_{1z}}\xi_{z} + \sqrt{\hat{v}_{1z}}\eta_z)z
        + \beta c \log p_{y|z}(y|z)
    }dz.
\end{eqnarray}
In the limit $l\to0, \beta\to\infty$, the extreme condition yields the same form of the equations that appear in the fixed point condition of the SE equations (\ref{eq:se_moment_matching_f_to_g_1})-(\ref{eq:se_moment_matching_g_to_f_4}). Additionally, at the extremum, the variational parameters $q_{x}, \chi_{x}, v_x$ and $m_x$ are in accordance with the right-hand side of the equations (\ref{eq:approx_qx})-(\ref{eq:approx_mx}). Similar accordance also holds for $q_z, \chi_z, v_z$ and $m_z$. Thus, the fixed point of SE of SA rVAMP is consistent with the replica symmetric calculation.

\section{Application to logistic regression}
\label{sec:numerical experiment}
\begin{figure}[t]
     \centerline{
        \includegraphics[width=\linewidth]{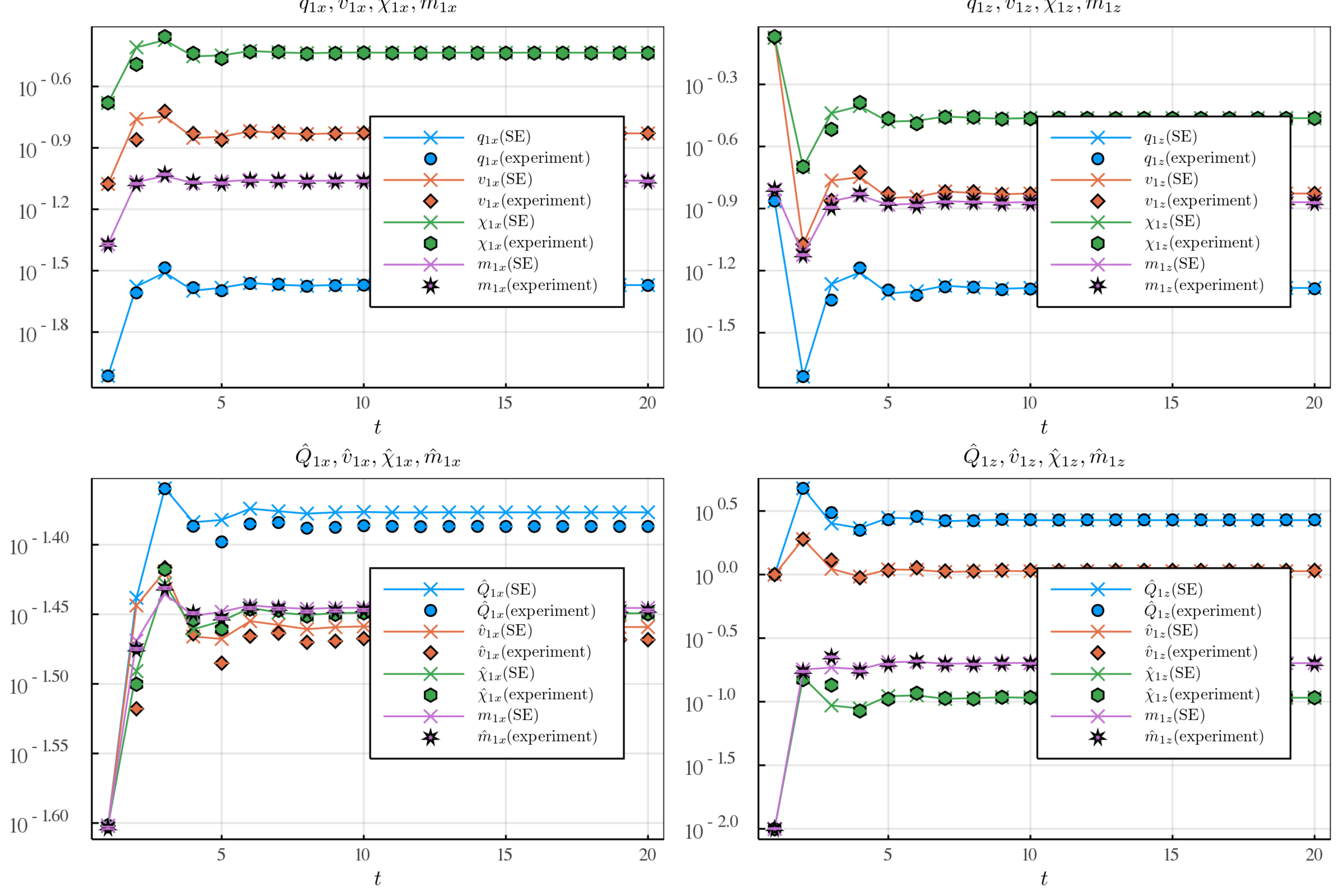}
     }
     \caption{
        Comparison between the iteration dynamics of SA rVAMP in the algorithm \ref{algo:self-averaging rVAMP} and in the SE equations defined in (\ref{eq:se_moment_matching_f_to_g_1})-(\ref{eq:se_moment_matching_g_to_f_4}).
        The solid lines show the SE trajectories. The symbols represent the median of SA rVAMP trajectories that are obtained from 1000 experiments.
        Top left: Macroscopic variables $q_{1x}^{(t)}, \chi_{1x}^{(t)}, v_{1x}^{(t)}, $ and $m_{1x}^{(t)}$ versus algorithm iteration. 
        Top right: Macroscopic variables $q_{1z}^{(t)}, \chi_{1z}^{(t)}, v_{1z}^{(t)}, $ and $m_{1z}^{(t)}$ versus algorithm iteration.
        Bottom left: Parameters $\hat{Q}_{1x}^{(t)}, \hat{v}_{1x}^{(t)}, \hat{\chi}_{1x}^{(t)}$ and $\hat{m}_{1x}^{(t)}$ versus algorithm iteration.
        Bottom right: Parameters $\hat{Q}_{1z}^{(t)}, \hat{v}_{1z}^{(t)}, \hat{\chi}_{1z}^{(t)}$ and $\hat{m}_{1z}^{(t)}$ versus algorithm iteration.
    }
    \label{fig:compare_SE}
\end{figure}
For checking the validity of the results obtained so far, we applied rVAMP to logistic regression and conducted numerical experiments in order to (i) validate our SE, (ii) obtain insights about the convergence speed from SE, and (iii) test the applicability of rVAMP to real-world problems.

In logistic regression, the domain of the response variables $\mathcal{Y}$ is $\{-1,1\}$, and the likelihood is given as 
\begin{equation}
    p_{y|z}(y|z) = \delta(y-1)\frac{1}{1 + e^{-z}} + \delta(y+1)\frac{1}{1+e^z}.
\end{equation}
Additionally, $g^\prime_{1z}$ in (\ref{eq:chi1z_susceptibility}) can be written as 
\begin{eqnarray}
\fl
    g_{1z}^\prime(h_{1z,\mu}, \hat{Q}_{1z,\mu}, \hat{v}_{1z,\mu}; c_\mu, \eta_{z,\mu}, y_\mu) = 
    \nonumber \\
    \left[
        \hat{Q}_{1z,\mu} + \frac{c_\mu}{4\cosh^2\left(
            \frac{1}{2}g_{1z}(h_{1z,\mu}, \hat{Q}_{1z,\mu}, \hat{v}_{1z,\mu}; c_\mu, \eta_{z,\mu}, y_\mu)
        \right)}
    \right]^{-1}.
\end{eqnarray}

All the experiments were conducted on a single Intel(R) Core(TM) i7-8700B (3.20GHz) CPU.

\begin{figure}[t]
     \centerline{
        \includegraphics[width=\linewidth]{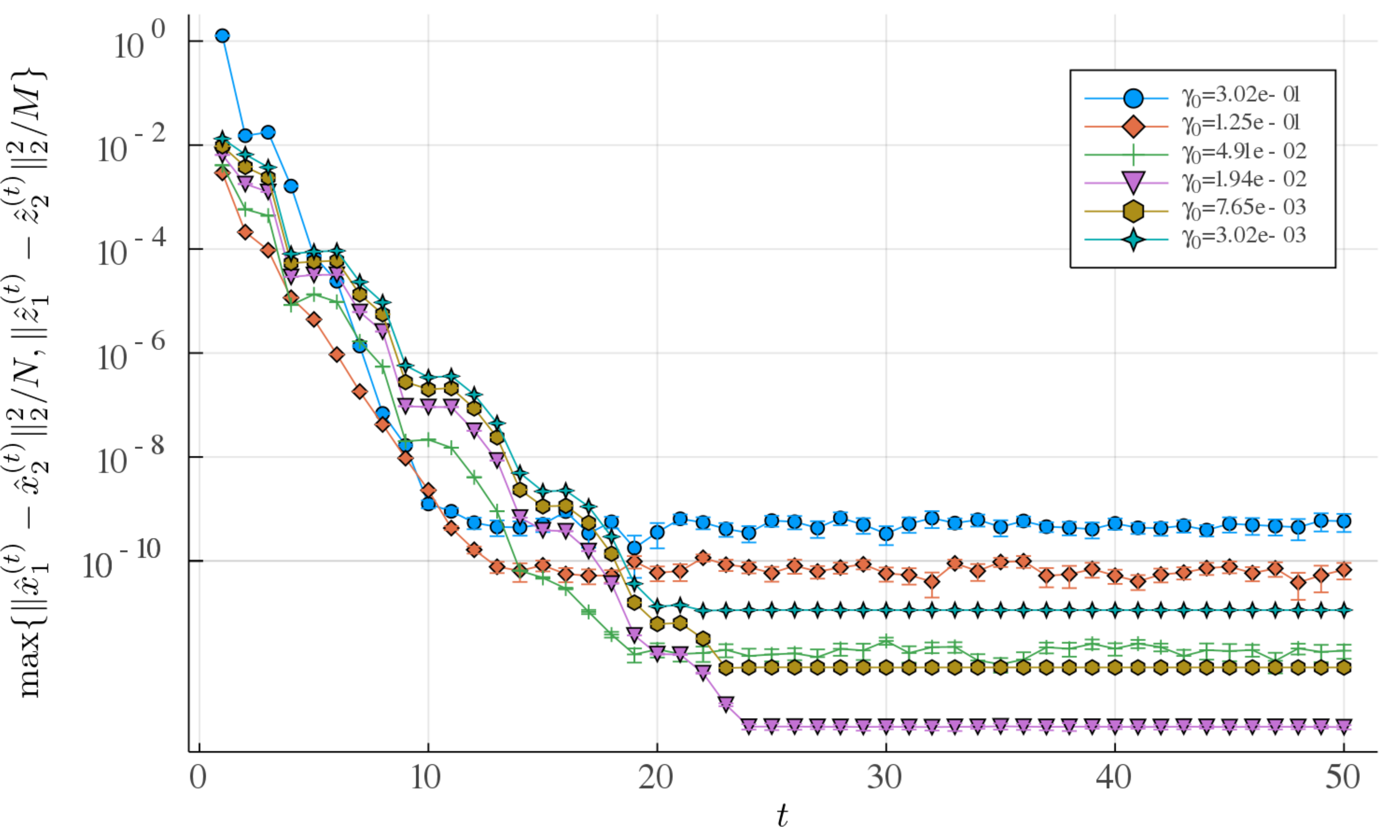}
     }
     \caption{
        Time evolution of the convergence criterion $\max\{\|\hat{\bm{x}}_1^{(t)} - \hat{\bm{x}}_2^{(t)}\|_2^2/N, \|\hat{\bm{z}}_1^{(t)} - \hat{\bm{z}}_2^{(t)}\|_2^2/M\}$ is plotted versus the iteration step $t$. The error bars represent the standard errors. The symbols represent the median of rVAMP trajectories obtained from 1000 experiments.
    }
    \label{fig:dynamics_aron_intercept}
\end{figure}

\subsection{Comparing with SE using synthetic data}

Synthetic data were generated under the settings described in subsection \ref{subsection:setup for macro analysis}. The actual data generation process are described by
\begin{eqnarray}
    q_{x_0}(x_{0,i})&= \rho \mathcal{N}(x_{0,i};0,\rho^{-1}) + (1-\rho)\delta(x_{0,i}), 
    \\
    q_{y|z}(y_\mu |\bm{a}_\mu^\top \bm{x}_0) &= \delta(y_\mu-1)\frac{1}{1 + e^{-\bm{a}_\mu^\top \bm{x}_0}} + \delta(y_\mu+1)\frac{1}{1+e^{\bm{a}_\mu^\top \bm{x}_0}},
\end{eqnarray}
where $\mathcal{N}(x_{0,i};\mu,\sigma^2)$ is the Gaussian measure with mean $\mu$ and variance $\sigma^2$, and $\rho\in[0,1]$ is the sparsity. The system size $N$, the measurement ratio $\alpha=M/N$, and the sparsity $\rho$ were specified as $N=10000, \alpha=0.2$, and $\rho=0.01$, respectively. Additionally, the feature matrix $A$ was drawn from the row-orthogonal ensemble \cite{kabashima2014signal} for which the limiting eigenvalue distribution of $A^\top A$ was $\rho(\lambda)=\alpha\delta(\lambda-1) + (1-\alpha)\delta(\lambda)$.

To validate SE, we compared the iteration dynamics of SA rVAMP to those of SE. Figure \ref{fig:compare_SE} plots the order parameters and the parameters of $p_1^{(\beta)}$ versus the iteration index $t$. The data of SA rVAMP were obtained from $1000$ random trials. The error bars are smaller than the size of the markers. Although some systematic disagreements are present in $\hat{Q}_{1x}^{(t)}$ and $\hat{v}_{1x}^{(t)}$ possibly due to the finite-size effect, most of the experimental values are in good agreement with the predictions of SE. This shows the validity of our SE.

The iteration dynamics of SE suggest that rVAMP converges in a few dozens of iterations, guaranteeing the fast convergence of rVAMP for the synthetic data.

\begin{figure}[t]
     \centerline{
        \includegraphics[width=\linewidth]{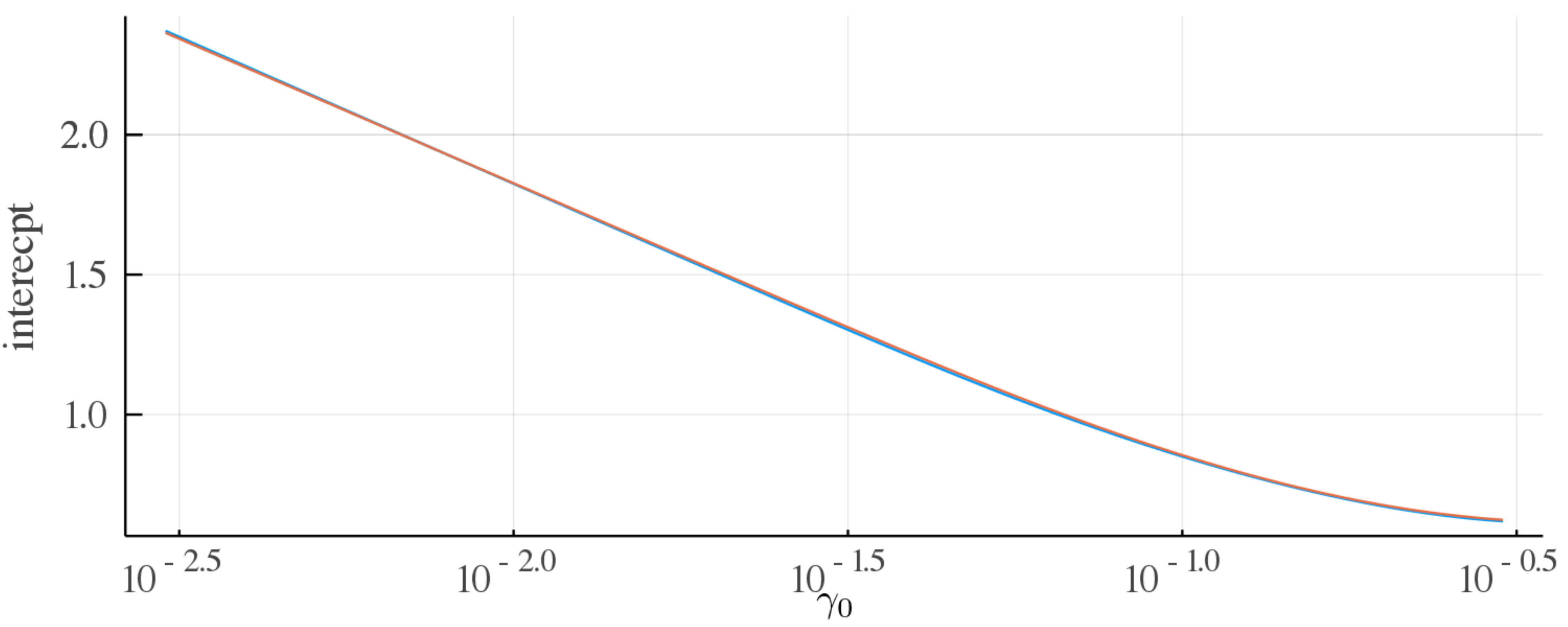}
     }
     \caption{
        Intercept term of logistic regression model plotted versus $\gamma_0$. The red line is obtained by the naive refitting procedure, while the blue line is obtained using rVAMP.
    }
    \label{fig:intercept term comparison}
\end{figure}

\subsection{Applicability of rVAMP in real world data}

\begin{figure}[t]
     \centerline{
        \includegraphics[width=\linewidth]{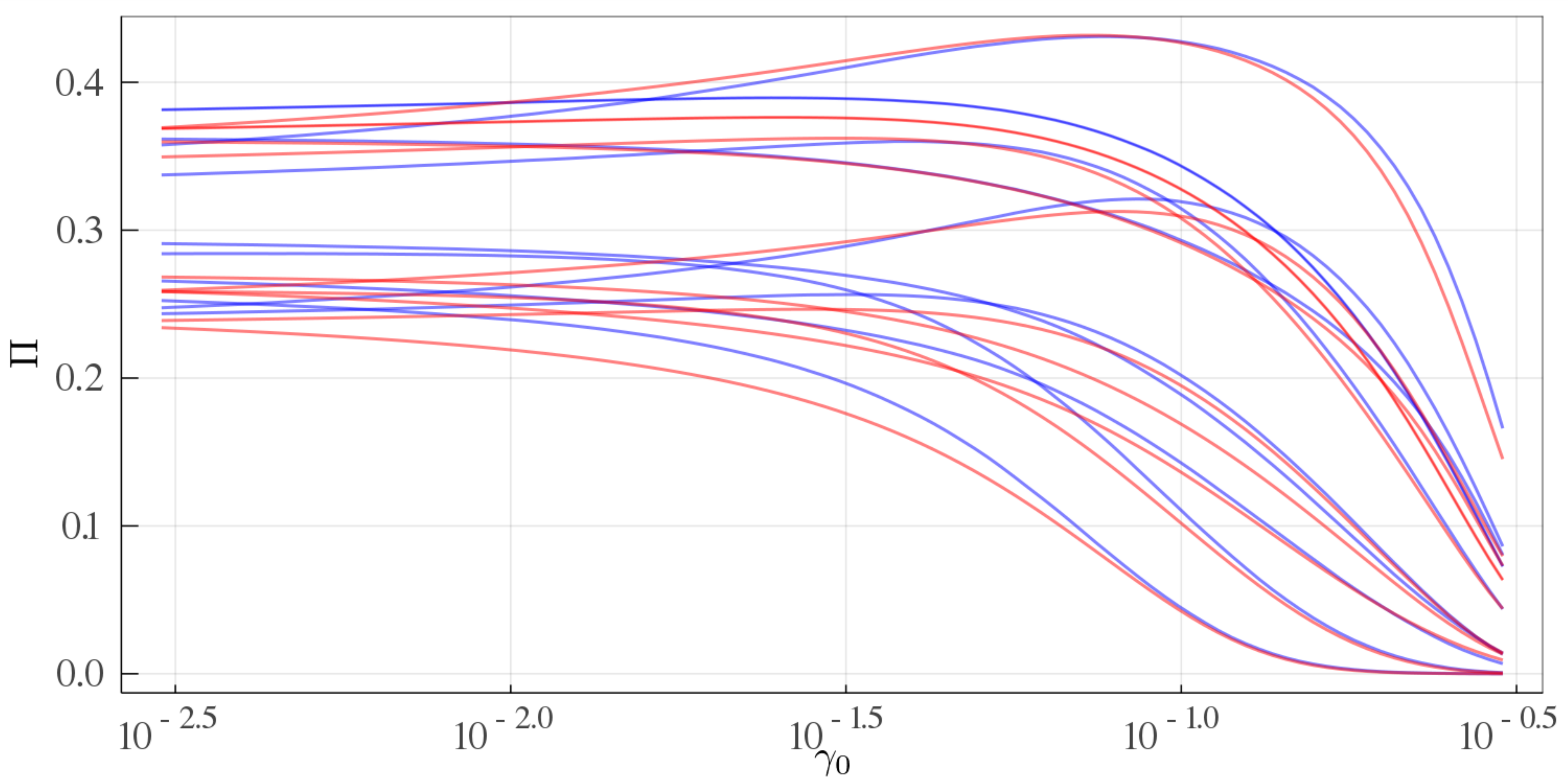}
     }
     \caption{
        Comparison of the selection probability plotted for various values of the regularization strength $\gamma_0$. For ease of viewing, the selection probabilities are shown  only for 10 features that had the largest selection probability for the smallest $\gamma_0$. Red lines are obtained using the naive refitting procedure, whle blue lines are obtained using rVAMP.
    }
    \label{fig:stabiltypath comparison}
\end{figure}

\begin{figure}[t]
     \centerline{
        \includegraphics[width=\linewidth]{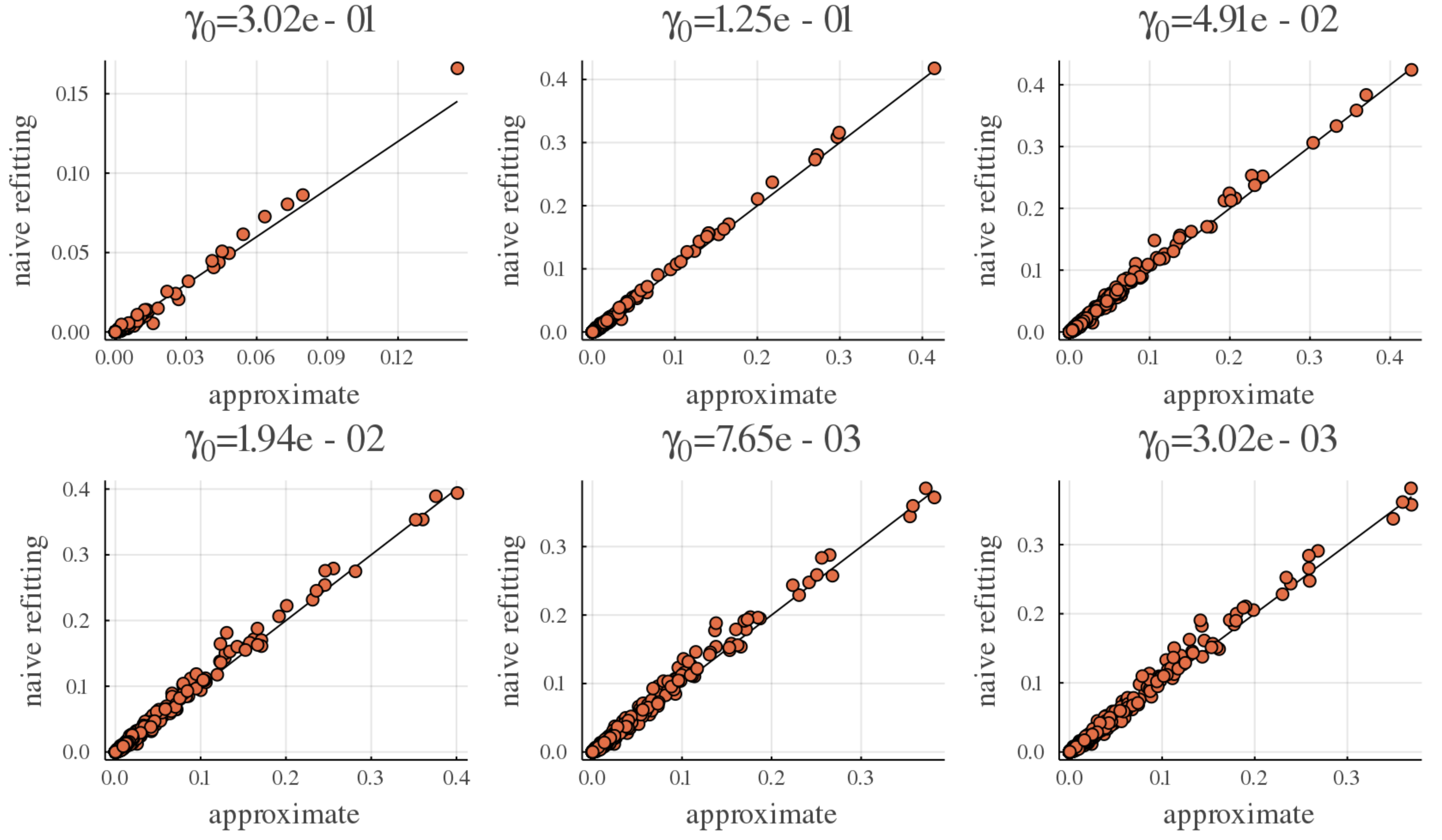}
     }
     \caption{
        Naive refitting estimates of the selection probability $\Pi_i,i=1,2,\dots,N$ plotted versus those computed by rVAMP for various regularization strengths.
    }
    \label{fig:stability selected}
\end{figure}

We explored the performance of rVAMP on the colon cancer dataset \cite{alon1999broad}, which is also used in the introduction. The data is publicly available at \url{http://genomics-pubs.princeton.edu/oncology/}. The task is to distinguish cancer from normal tissues using micro-array data with $N = 2000$ features per example. The data
were derived from 22 normal ($y_\mu=-1$) and 40 ($y_\mu=1$) cancer tissues. The total
number of samples is $M = 62$. We pre-processed the data by carrying out base $10$ logarithmic transformation and standardizing each feature to zero mean and unit variance. Because the class labels are biased, we included the intercept term. To obtain the selection probabilities for a grid of $\gamma_0$, we used the warm start procedure. Finally, the damping factor $\eta_{\rm d}$ was set to $0.85$.

First, we examined the convergence speed of rVAMP. Figure \ref{fig:dynamics_aron_intercept} shows the time evolution of the convergence criterion $\max\{\|\hat{\bm{x}}_1^{(t)} - \hat{\bm{x}}_2^{(t)}\|_2^2/N, \|\hat{\bm{z}}_1^{(t)} - \hat{\bm{z}}_2^{(t)}\|_2^2/M\}$ by plotting its value versus the iteration step $t$. For various regularization strengths, regular exponential decay is observed, This demonstrating the fast convergence of rVAMP in a real-world dataset.

\begin{figure}[t]
     \centerline{
        \includegraphics[width=\linewidth]{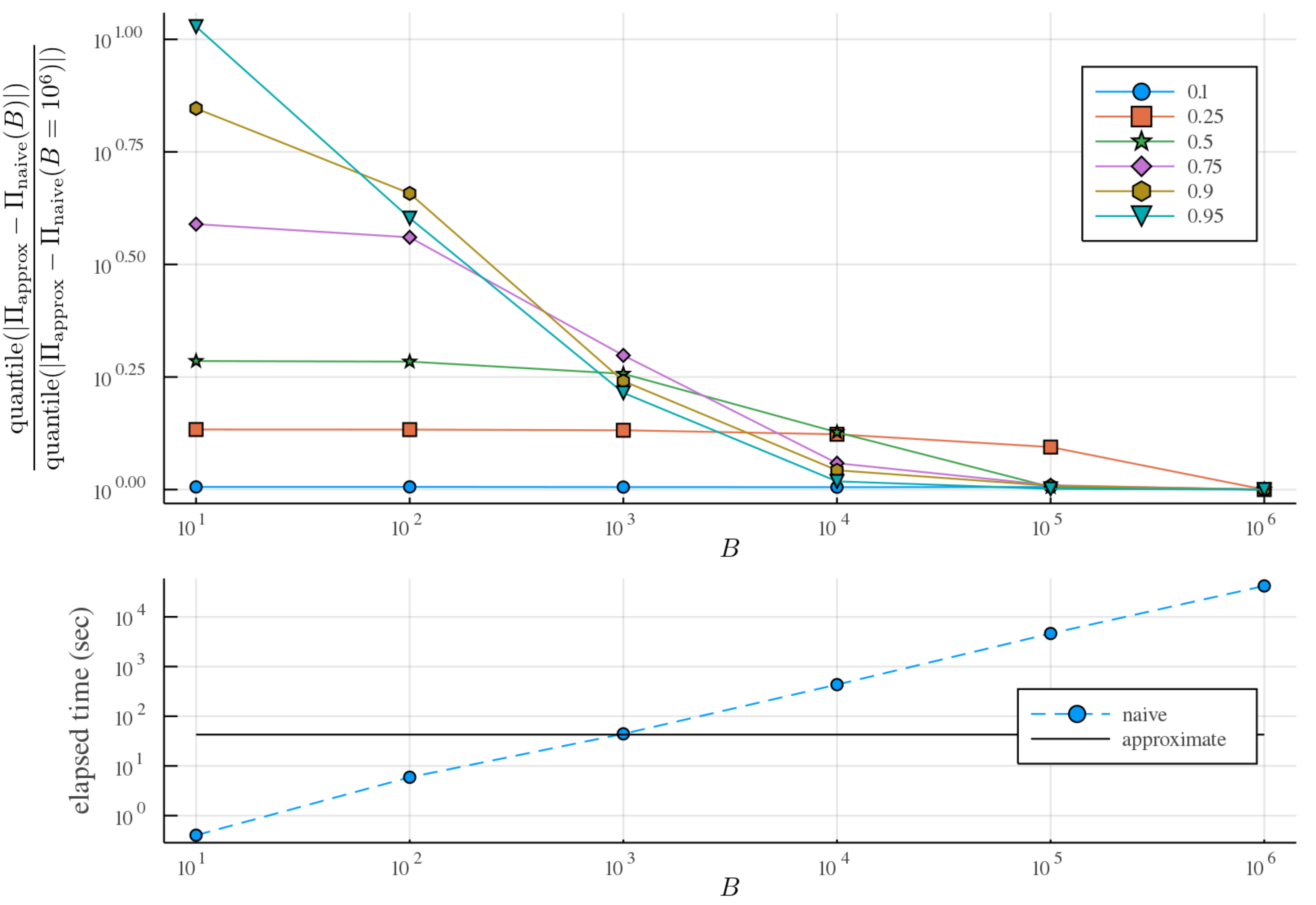}
     }
     \caption{
        Upper panel: The difference between approximated and naively calculated selection probabilities  plotted versus the number of resampled datasets $B$. We denote by $\Pi_{\rm approximate}$ the selection probability obtained by rVAMP, and by $\Pi_{\rm naive}$ that obtained by naive resampling procedure using $B$ resampled datasets. The difference is measured as a $q$-quantile of the difference for all of the selection probabilities in the grid of $\gamma_0$.
        Lower panel: Elapsed time is plotted versus the size of the resampled dataset $B$.    }
    \label{fig:accuracy and elapsed time}
\end{figure}

Next, we examine the accuracy of rVAMP. To compare the estimate of rVAMP with that of the naive refitting procedure of SS, the naive refitting on 1,000,000 resampled datasets was conducted using GLMNet \cite{glmnetmatlab}.
Figure \ref{fig:intercept term comparison} shows the intercept term plotted versus the regularization strength. For a wide range of $\gamma_0$, rVAMP accurately estimated the intercept term.  Figure \ref{fig:stabiltypath comparison} plots the comparison between the selection probabilities estimated by rVAMP and by the naive reffiting for the entire grid of $\gamma_0$. For ease of viewing, we only plot these values for the $10$ features that had the largest selection probabilities for the smallest $\gamma_0$. Figure \ref{fig:stability selected} plots the same comparison of all of the features for a selected set of $\gamma_0$. Although the accuracy decreases slightly as we weaken the regularization, rVAMP successfully approximate the selection probability. The upper panel of ignore \ref{fig:accuracy and elapsed time} plots the difference between approximated and naively calculated selection probabilities as a function of the number of resampled datasets $B$.  These results also provide evidence for the accuracy of rVAMP. The lower panel of figure \ref{fig:accuracy and elapsed time} plots the elapsed time used to obtain all of the selection probabilities for various $\gamma_0$. Although the actual computation time depends on the implementation, this figure suggests that rVAMP can provide accurate estimate of $\bm{\Pi}$ in a much shorter time than the naive SS. These observations demonstrate the accuracy of rVAMP.

\section{Summary and conclusion}
\label{sec:summary and conclusion}
We developed an approximate SS algorithm that enables SS without the use of the repeated fitting procedure. The key concept is to use the combination of the replica method of statistical mechanics and the VAMP algorithm of information theory. The derivation of the algorithm was based on the expectation propagation of machine learning. We also derived the state evolution that macroscopically describes the dynamics of the proposed algorithm, and showed that its fixed point is consistent with the replica symmetric solution.  Through numerical experiments, we confirmed that the state evolution equation is valid and that the proposed algorithm converges in a few dozens of iterations. We applied the proposed algorithm to logistic regression and demonstrated its application to a real-world dataset through numerical experiments. Although the real-world dataset has statistical correlations among the features, the proposed algorithm achieved fast convergence and high-estimation accuracy, demonstrating its utility for real-world problems.

A possible drawback of our algorithm is its computational complexity, even though it was not significant for the experiments described in section \ref{sec:numerical experiment}. Because the algorithm requires the computation of matrix inversion at each iteration, the computational burden may increase significantly with the increasing number of samples in the datasets. This shortcoming may be addressed by the self-averaging version of the proposed algorithm or the dual-decomposition-like variable augmentation used in the alternating direction method of multipliers \cite{boyd2011distributed, boyd2004convex}. 

A promising future research direction includes analyzing the variable selection performance of the SS algorithm using SE. Generally, theoretical analysis of resampling techniques is difficult in general because we cannot explicitly write down the analytical form of the estimators. This difficulty prevents the obtaining of useful insights from quantitative theoretical analysis. Thus, the replica theory \cite{mezard1987spin} may provide a promising analytical tool in this area. Because our framework can treat only synthetic settings, we believe that the goal is to investigate precise asymptotic properties for a comprehensive range of parameters and to find some phenomena that would hold universally, such as novel phase transitions. However, this kind of exhaustive analysis is quite involving in practice, although obtaining an order parameter for one specific setting is not difficult. Thus we postpone this analysis as future work. Another research direction is the investigation of the dynamics of raw rVAMP using techniques such as the dynamical-functional theory \cite{ccakmak2019memory, cakmak2017dynamical, martin1973statistical, eissfeller1992new}.

\section*{Acknowledgement}
This work was supported by JSPS KAKENHI Grant Numbers 19J10711, 17H00764, and JST CREST Grant Number JPMJCR1912, Japan.

\section*{References}
\bibliographystyle{iopart-num}
\bibliography{main}
\end{document}